\documentclass[PCfour,sageh,times]{sagej}

\usepackage{moreverb,url}
\usepackage[colorlinks,bookmarksopen,bookmarksnumbered,citecolor=blue,urlcolor=blue,linkcolor=blue]{hyperref}

\usepackage{amsmath}
\usepackage{amssymb}
\usepackage{graphicx}
\usepackage{booktabs}
\usepackage{array}
\usepackage{algorithm}
\usepackage{algpseudocode}
\algrenewcommand{\algorithmiccomment}[1]{\hfill\(\triangleright\)\ \textit{#1}}

\usepackage{pgfplots}
\pgfplotsset{compat=1.18}
\usepgfplotslibrary{groupplots}
\definecolor{rlblue}{RGB}{31,90,158}
\definecolor{rlred}{RGB}{178,34,52}

\graphicspath{{./}{../}}

\usepackage{xcolor}
\usepackage{orcidlink}
\definecolor{todoblue}{RGB}{0,87,217}
\newif\ifdraftnotes
\draftnotestrue

\providecommand{\R}{\mathbb{R}}
\providecommand{\E}{\mathbb{E}}
\def\eqref#1{equation~\ref{#1}}
\def\plaineqref#1{\ref{#1}}

\def\volumeyear{2026}

\makeatletter
\renewcommand\subsection{\@startsection{subsection}{2}{\z@}%
    {0.9\@bls plus .3\@bls minus .1\@bls}%
    {4pt\@afterindentfalse}%
    {\sagesf\normalsize\bfseries\raggedright}}
\renewcommand\subsubsection{\@startsection{subsubsection}{3}{\z@}%
    {0.5\@bls plus .3\@bls minus .1\@bls}%
    {-0.5em\@afterindentfalse}%
    {\sagesf\normalsize\bfseries}}
\makeatother

\begin{document}

\runninghead{Zhu et al.}

\title{Towards High-DoF Dexterous Manipulation through VLA Post-Training}

\author{
  Junlei Zhu\affilnum{1,2}\textsuperscript{*}\orcidlink{0009-0006-6518-3575},
  Shenzhe Yao\affilnum{1}\textsuperscript{*}\orcidlink{0009-0006-0016-5501},
  Chaogui Huang\affilnum{1}\orcidlink{0009-0007-3097-1667},
  Wenkai Zhu\affilnum{1,2}\orcidlink{0009-0002-1928-5934}, 
  Jingwei Peng\affilnum{1,2}\orcidlink{0009-0006-9428-6004}, Guanqi He\affilnum{1}\orcidlink{0009-0007-6016-9800},
  S{\"o}ren Schwertfeger\affilnum{2}\orcidlink{0000-0003-2879-1636},
  Jiahao Chen\affilnum{1}\orcidlink{0000-0002-8927-5646} and
  Yide Liu\affilnum{1}\orcidlink{0000-0002-2447-3107}
}

\affiliation{\affilnum{1}Wuji Technology\\
\affilnum{2}School of Information Science and Technology, ShanghaiTech University, China\\
\textsuperscript{*}These authors contributed equally to this work.}

\corrauth{Yide Liu, Wuji Technology.}

\email{liu.yide@wuji.tech}

\begin{abstract}
Imitation-learned vision--language--action (VLA) foundation models acquire broad manipulation capabilities by scaling robot data across tasks and embodiments, but reliable deployment on a specific downstream task and hardware platform still requires post-training.
Dexterous hands make this adaptation particularly difficult: their broad behavioural repertoire and high degree of freedom create a large and structured action space.
Three obstacles are central: open-source VLAs do not natively provide an action interface for high-DoF hands; gesture mismatch during human-gated DAgger takeover creates command discontinuities and contaminates corrective trajectories; and reinforcement learning in the raw joint space is sample-inefficient.
We present a unified four-step post-training pipeline comprising a learned temporal hand-action codec, supervised fine-tuning, DAgger, and real-world residual reinforcement learning.
The codec adapts a pretrained VLA to absolute dexterous-hand commands.
Buffered rollback, pose alignment, and smooth command blending enable continuous, task-relevant DAgger corrections, while latent residual RL confines exploration to coordinated hand motions captured by the codec.
We evaluate the pipeline on five diverse real-world tasks spanning bimanual transfer, in-hand reorientation, and tool use.
Within the reported post-training budgets, the resulting policies achieve 100\% success on every evaluated task over 20 trials per task.
These results provide a practical path for adapting VLA foundation models to reliable real-world dexterous manipulation.
\end{abstract}


\maketitle
\footnotetext[0]{\sagesf Project website: \href{https://wuji.tech/blog/post-training-0}{wuji.tech/blog/post-training-0}.}

\begin{figure*}[t!]
\centering
\includegraphics[width=\textwidth]{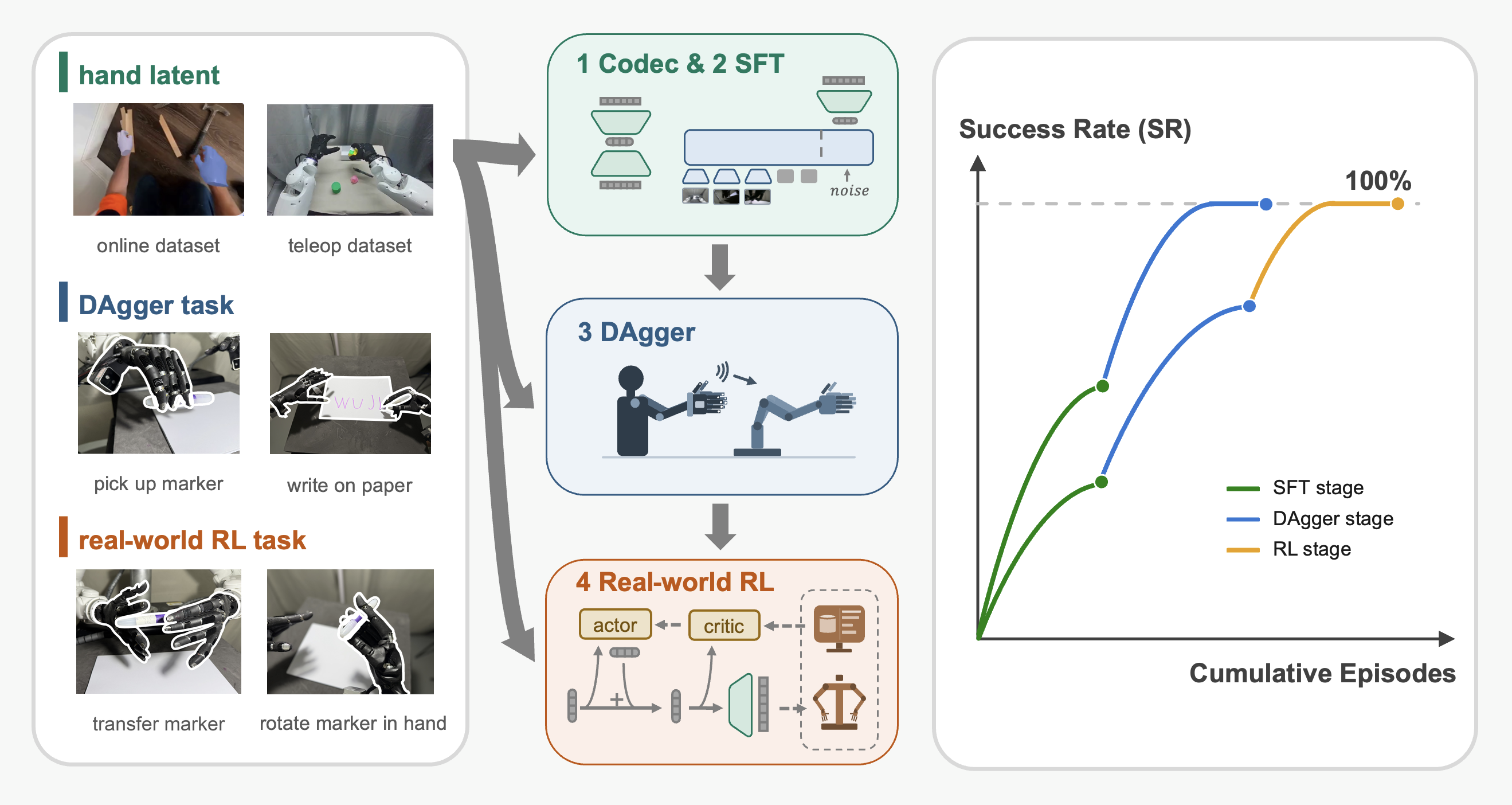}
\caption{Overview of dexterous policy post-training.
A learned hand-action representation provides a common interface for supervised fine-tuning, corrective data aggregation, and real-world latent residual reinforcement learning, which progressively improve task success.
The curves on the right schematically illustrate improvement across post-training stages.}
\label{fig:teaser}
\end{figure*}

\begin{table*}[t!]
\caption{Important symbols used throughout the paper, grouped by pipeline stage.}
\label{tab:notation}
\centering
\footnotesize
\setlength{\tabcolsep}{3pt}
\renewcommand{\arraystretch}{1.12}
\begin{tabular}{@{}
  >{\raggedright\arraybackslash}p{2.60cm}
  >{\raggedright\arraybackslash}p{3.95cm}
  @{\hspace{2.5mm}}
  >{\raggedright\arraybackslash}p{2.10cm}
  >{\raggedright\arraybackslash}p{4.45cm}@{}}
\toprule
\textbf{Symbol} & \textbf{Brief description} &
\textbf{Symbol} & \textbf{Brief description} \\
\midrule
\multicolumn{4}{@{}l}{\textit{Step 1: Hand-action codec}} \\
\addlinespace[1pt]
$H$ & Action-chunk horizon ($H=32$) &
$E^{\mathrm{hand}},D^{\mathrm{hand}}$ & Per-hand codec encoder and decoder \\
$q,q_0\in\R^{54}$ & Current state; chunk-start state &
$E,D$ & Full action-codec encoder and decoder \\
$q^{\mathrm{hand}}_0\in\R^{20}$ & Hand state at chunk start &
$z^{\mathrm{hand}}_{1:H}\in\R^{H\times9}$ & Per-hand latent chunk \\
$q^{\mathrm{hand,cmd}}_{1:H}\in\R^{H\times20}$ & Commanded hand-joint chunk &
$z_{1:H}\in\R^{H\times32}$ & Full latent-action chunk \\
$q^{\mathrm{cmd}}_t\in\R^{54}$ & Executable joint command &
$\mu,\sigma,\epsilon$ & VAE mean, std., and sampling noise \\
$a_{1:H}\in\R^{H\times54}$ & Joint-space action chunk &
$\Delta^k,e_{t,j}^{(k)}$ & Temporal-difference operator; error \\
$a^{\mathrm{hand}}_{1:H}\in\R^{H\times20}$ & Relative hand-action chunk &
$\tau_{\mathrm{Huber}},\beta_{\mathrm{KL}}$ & Huber threshold and KL weight \\
$\widehat a^{\mathrm{hand}}_{1:H}\in\R^{H\times20}$ & Reconstructed hand-action chunk &
$w_j,\lambda_k$ & Joint and temporal loss weights \\
\midrule
\multicolumn{4}{@{}l}{\textit{Step 2: Supervised fine-tuning}} \\
\addlinespace[1pt]
$C=25$ & Executed frames per policy query &
$o=(I_{1:K},\ell,q)$ & Observation: views, language and state \\
$t_{\mathrm{flow}}\in[0,1]$ & Flow time &
$x_{t_{\mathrm{flow}}}\in\R^{H\times32}$ & Interpolated latent \\
$\widehat z\in\R^{H\times32}$ & One-step clean-latent estimate &
$v_\omega,u$ & Predicted vector field and flow target \\
$\lambda_{\mathrm{arm}},\lambda_{\mathrm{lat}},\lambda_{\mathrm{dec}}$ & SFT loss weights &
& \\
\midrule
\multicolumn{4}{@{}l}{\textit{Step 3: DAgger}} \\
\addlinespace[1pt]
$q^{\mathrm{hold}}\in\R^{20}$ & Pre-takeover hand command &
$q^{\mathrm{teleop}}_t\in\R^{20}$ & Live operator command \\
$s$ & Elapsed blend time &
$\tau_b$ & Blend duration \\
$\alpha_s$ & Blend fraction at elapsed time $s$ &
$\alpha_{\mathrm{new}}$ & New-correction batch fraction \\
\midrule
\multicolumn{4}{@{}l}{\textit{Step 4: Real-world residual RL}} \\
\addlinespace[1pt]
$\pi_{\mathrm{ref}}$ & Frozen reference policy &
$\mathcal B$ & Replay buffer \\
$z_{\mathrm{ref}}\in\R^{H\times32}$ & Reference-policy latent chunk &
$r_t,d_t,\gamma$ & Reward, terminal flag and discount \\
$\eta_{1:M}\in\R^{M\times2048}$ & Prefix image-token features &
$Q_{\phi_i},Q_{\bar\phi_i}$ & Twin critics and targets \\
$\xi_{\mathrm{rl}}\in\R^{2048}$ & Learnable readout token &
$\tau_{\mathrm{Polyak}}$ & Target averaging coefficient \\
$g_\varphi,W_{\mathrm{rl}}$ & Readout transformer/projection &
$\mathcal R_{\Delta z},\mathcal S$ & Residual/smoothness penalties \\
$h_{\mathrm{rl}}\in\R^{2048}$ & RL-token observation representation &
$\beta_q,\beta_{\Delta z},\beta_s$ & Actor-loss weights \\
$f_\theta,f_{\bar\theta}$ & Actor and TD3 target actor &
$b^c,b^z$ & Correction and latent bounds \\
$c_{1:C},\tilde c_{1:C}\in\R^{C\times32}$ & Bounded and noise-perturbed corrections &
$\sigma^{\mathrm{arm}},\sigma^{\mathrm{hand}}$ & Exploration noise scales \\
$\Delta z\in\R^{H\times32}$ & Full-horizon latent residual &
$\nu$ & Noise spectral exponent ($\nu=2$) \\
$m_{\mathrm{task}}\in\{0,1\}^{32}$ & Binary residual mask &
$\dot q_{\max}$ & Maximum command change per frame \\
\bottomrule
\end{tabular}
\end{table*}

\section{Introduction}
\label{sec:intro}

Vision--language--action (VLA) foundation policies are increasingly capable across tasks and embodiments \citep{brohan2023rt2,openx2024rtx,octo2024,black2024pi0,intelligence2025pi05}.
Despite this progress, deployment on a particular downstream task and hardware platform can still expose systematic failures caused by differences in task distribution, embodiment, sensing, action interfaces, and encountered states.
Task- and robot-specific post-training is therefore needed to turn these broad capabilities into reliable execution.

Dexterous manipulation makes this adaptation particularly challenging.
Compared with parallel grippers, dexterous hands support a broader range of manipulation behaviors, expose many more controlled degrees of freedom, and require coordinated finger postures under multi-contact dynamics \citep{rajeswaran2018dexterous,andrychowicz2020dexterous}.
The resulting action space is both high-dimensional and strongly structured.
This creates three obstacles across the post-training pipeline: establishing a compatible action interface, collecting clean human corrections, and exploring useful policy improvements efficiently on the real robot.

\textbf{The first obstacle is the mismatch between the pretrained VLA's action interface and high-DoF hand control.}
The pretrained policy's fixed action output does not directly match the joint-command space of our bimanual dexterous embodiment.
We therefore train a temporal hand codec that maps coordinated hand trajectories into a compact latent action space compatible with the VLA.

\textbf{The second obstacle is gesture mismatch during DAgger intervention, which causes discontinuous high-DoF hand commands and contaminates corrective supervision.}
Recent dexterous human-in-the-loop systems have identified intervention discontinuity in high-DoF control, and HandITL attributes it to command mismatch between the operator and the policy and avoids it through relative hand retargeting and shared arm control \citep{han2026dexhil,li2026handitl}.
We study a different interface constraint: both teleoperation and the reference VLA use absolute hand commands, which we retain so that demonstrations, corrections, and policy outputs share the same action semantics.
Under this interface, directly switching to teleoperation when the operator's live gesture differs from the robot hand's current commanded pose produces a discontinuous joint command.
Beyond the physical discontinuity, the initial recorded motion may describe gesture realignment rather than recovery from the policy failure, contaminating the corrective supervision.
We therefore separate pose synchronization from task recovery by rolling the robot back to a buffered pre-failure state, displaying the operator and robot hand poses for alignment, and blending the hand command over $2$\,s at the control switch.
This mechanism preserves absolute hand control while producing continuous, task-relevant correction trajectories.

\textbf{The third obstacle is sample-inefficient exploration in the high-dimensional raw joint space of a dexterous hand.}
As the number of independently controlled joints increases, online reinforcement learning must search over a rapidly expanding space of possible corrections, making effective behaviors difficult to discover from limited interaction \citep{rajeswaran2018dexterous}.
This problem is compounded by contact-rich dynamics, for which simulation is difficult to model accurately and real-robot samples are expensive \citep{andrychowicz2020dexterous,akkaya2019rubik,xu2026rltoken}.
Consequently, dexterous post-training calls for an off-policy method that can reuse collected experience while making each online correction structurally meaningful \citep{lillicrap2016ddpg,fujimoto2018td3}.
We meet this requirement with latent residual reinforcement learning, which searches for corrections in a compact hand-action representation learned from coordinated motion rather than directly perturbing individual joint coordinates \citep{santello1998postural,he2021synergies}.

Our goal is to turn a pretrained VLA into a reliable policy for a specified real-world dexterous skill.
We focus on atomic tasks that can be completed in approximately $30$\,s; long-horizon task composition and scene-level generalization are outside the scope of this work.
To this end, we introduce a four-step post-training pipeline.
We first train and freeze a temporal codec that represents coordinated hand motion in a compact latent action space \citep{santello1998postural,kingma2014vae,he2021synergies}.
We then fine-tune the VLA with supervised learning in this space, aggregate human corrections through DAgger, and finally freeze the corrected policy and learn a latent residual through online reinforcement learning \citep{ross2011dagger,johannink2019residual,xu2026rltoken}.
Together, these stages provide a progression from task initialization with demonstrations, through correction on policy-induced states, to reward-driven improvement from real-world interaction.

We evaluate the pipeline on five real-world marker-manipulation tasks spanning grasping, bimanual transfer, in-hand rotation, object separation, and writing.
Starting from 200 demonstrations per task, we apply DAgger and latent residual RL only when the preceding stage leaves systematic failures.
The successive stages raise every task to 100\% success over 20 evaluation trials.

We make four contributions.
\begin{itemize}
\item First, we present an end-to-end post-training pipeline that integrates SFT, DAgger, and latent residual RL through a shared hand-action representation for real-world dexterous skills.
\item Second, we formulate the DAgger handoff problem under an absolute hand-command interface and identify two coupled consequences of gesture mismatch: executed-command discontinuity and contamination of corrective supervision by gesture-realignment motion.
\item Third, we formulate latent residual RL that refines the actions of a frozen VLA through corrections in the learned hand-action space.
\item Fourth, we conduct extensive real-robot experiments that evaluate the complete pipeline, quantify the gains obtained at successive post-training stages, compare DAgger corrections with additional demonstrations, and compare latent with raw joint-space residual learning under matched budgets.
\end{itemize}

Videos of the real-robot results are available on the \href{https://wuji.tech/blog/post-training-0}{project website}.

\section{Related Work}
\label{sec:related}

\subsection{Foundation policies and dexterous action representations}

Large-scale robot datasets and vision--language--action (VLA) models have produced increasingly capable policies that transfer across tasks, embodiments, and semantic instructions \citep{brohan2023rt2,openx2024rtx,octo2024,black2024pi0,intelligence2025pi05}.
Task- and robot-specific deployment nevertheless commonly begins with supervised fine-tuning on demonstrations.
Dexterous manipulation makes this adaptation especially difficult because useful hand actions are high-dimensional but strongly coordinated.
Biomechanical studies identify low-dimensional postural synergies in human grasps \citep{santello1998postural}, and robotic methods learn related action synergies across manipulation tasks \citep{he2021synergies}.
Human-video pretraining pipelines provide a further source of structured hand-motion supervision for robot-policy training \citep{li2026vitra}.

Earlier dexterous-learning systems combine demonstrations with reinforcement learning or use large-scale simulation and domain randomization to acquire task-specific hand skills \citep{rajeswaran2018dexterous,andrychowicz2020dexterous,akkaya2019rubik}.
These results establish the feasibility of learned dexterity, but do not provide a shared post-training action representation for a pretrained VLA.
Our temporal hand codec is trained once and then frozen, providing a common action interface for supervised adaptation, interactive correction, and real-world reinforcement learning.

\subsection{Interactive correction for dexterous policies}

DAgger addresses compounding imitation-learning error by collecting expert labels under the learner-induced state distribution \citep{ross2011dagger}, while human-gated variants concentrate supervision on states at which an operator elects to intervene \citep{kelly2019hgdagger}.
For real-robot deployment, Sirius uses intervention events to reweight mixed human and robot experience during continual policy learning \citep{liu2025sirius}.
Recent dexterous VLA systems bring this paradigm to coordinated arm--hand control.
DexHiL identifies intervention discontinuity as a challenge in high-DoF dexterous control and combines an integrated teleoperation interface with intervention-aware sampling \citep{han2026dexhil}.
HandITL further attributes gesture jumps at takeover to command mismatch between human teleoperation and policy execution, and avoids them through relative hand retargeting and shared arm control \citep{li2026handitl}.

We retain the absolute hand-command interface shared by teleoperation and the reference VLA.
At takeover, a mismatch between the policy and operator commands can both produce a discontinuity in the executed command and introduce pose-synchronization motion into the correction data.
We address these effects through buffered rollback, pre-takeover pose alignment, and a $2$-s blend of the executed hand command, without changing the action semantics.

\subsection{Frozen-policy reinforcement learning and latent action spaces}

Residual reinforcement learning preserves an existing controller and learns a corrective action on top of it \citep{johannink2019residual}.
Recent VLA post-training systems use experience in several different ways.
$\pi^*_{0.6}$ and its RECAP procedure combine demonstrations, autonomous rollouts, and interventions through offline RL and advantage-conditioned policy learning \citep{amin2025pistar06}.
RL Token instead freezes the VLA, reads out a compact backbone feature, and trains small online actor--critic networks \citep{xu2026rltoken}.
BORA combines an offline action-conditioned critic with online action-residual adaptation for real-world dexterous VLAs \citep{chen2026bora}.

A complementary line of work asks where policy improvement should act.
PLAS constrains offline RL to the support of a learned action decoder \citep{zhou2021plas}, and LASER learns task-family action manifolds for more efficient simulated RL \citep{allshire2021laser}.
For frozen generative robot policies, DSRL steers the diffusion or flow noise latent \citep{wagenmaker2025dsrl}, whereas ZPRL perturbs an observation-side information bottleneck that conditions the frozen action generator \citep{yu2026zprl}.
Our residual acts instead in an action-side hand codec shared with the VLA's supervised training.
The reference policy remains frozen, a zero residual exactly recovers its decoded action, and the same real-robot budget can be used to compare latent and raw joint-space residuals.

\section{Method}
\label{sec:method}

\begin{figure*}[t]
\begin{center}
\includegraphics[width=0.92\textwidth]{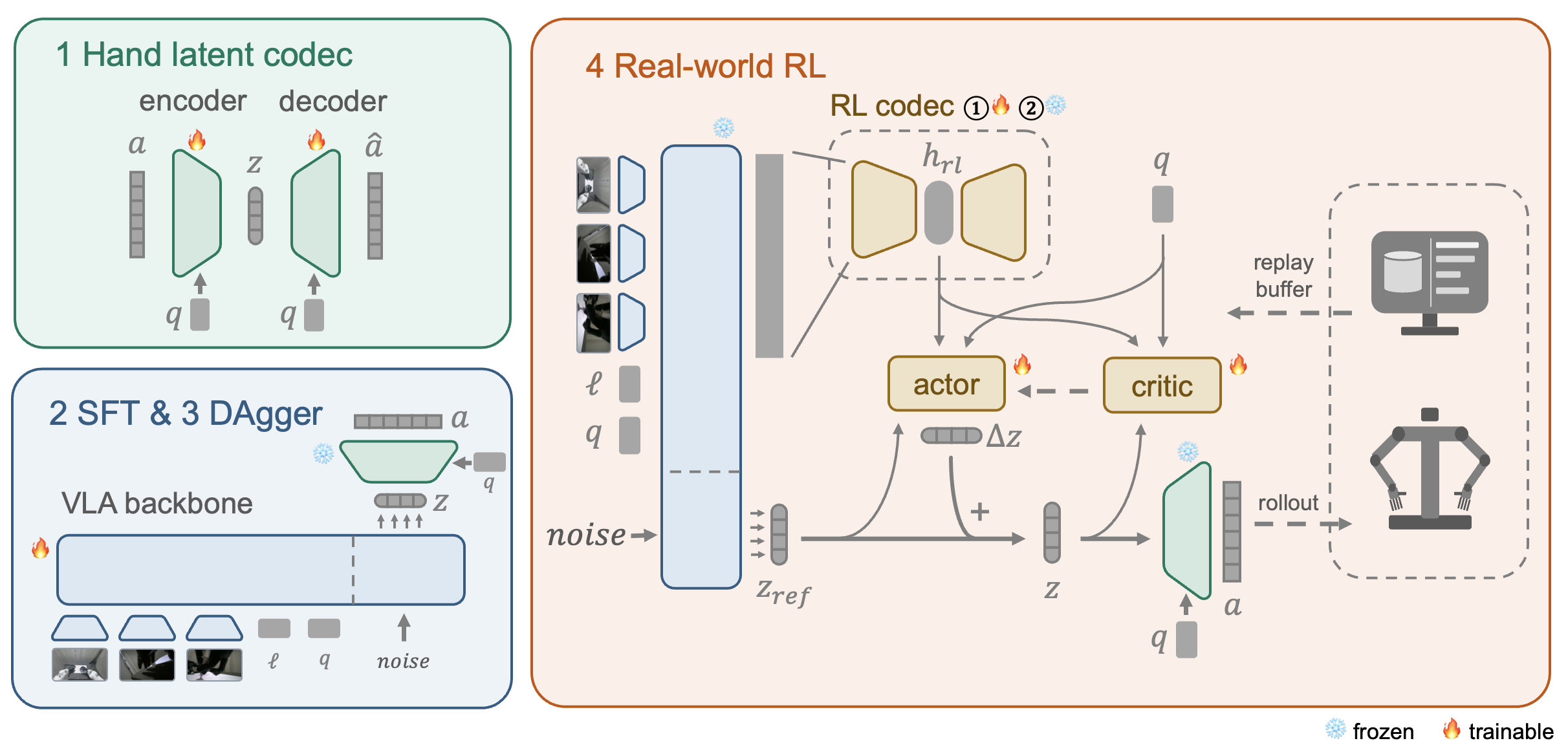}
\end{center}
\caption{Overview of our pipeline. \textbf{Codec.} A chunk-level VAE is trained
per hand on teleoperation, task and retargeted human-video data, compressing
each hand from 20 to 9 dimensions; the resulting codec $(E, D)$ is frozen.
\textbf{SFT and DAgger.} Demonstrations are expressed through $E$ as latent supervision
targets, and the VLA is fine-tuned to emit latent actions in that space,
with DAgger-style operator takeovers folded back into the corpus. The policy is
frozen thereafter. Here, $\ell$ denotes the language instruction and $q$ the
measured proprioceptive state. \textbf{Real-world RL.} The RL-token encoder is trained on
demonstration data before online RL \textcircled{1} and kept frozen during online RL \textcircled{2}.
Online, the frozen policy emits
$z_{\mathrm{ref}}$ with no encoding step, the RL token $h_{\mathrm{rl}}$ is read
out from the frozen prefix, and an actor conditioned on $(h_{\mathrm{rl}},q,z_{\mathrm{ref}})$
adds a bounded latent residual $\Delta z$ that is decoded once into
the executed chunk. Rollouts fill a replay buffer from which the critics and the
actor are trained.}
\label{fig:overview}
\end{figure*}

Our pipeline consists of four steps.
We first train a codec that compresses dexterous hand motion into a low-dimensional latent, and validate that its decoder faithfully reconstructs feasible hand motion (\hyperref[sec:method:latent]{Step~1}).
We then fine-tune a vision-language-action policy that acts in that latent space (\hyperref[sec:method:sft]{Step~2}).
Next, we refine the policy through corrective data aggregation (\hyperref[sec:method:dagger]{Step~3}).
Finally, we freeze that policy and use online residual reinforcement learning (RL) to learn a correction confined to the latent action space (\hyperref[sec:method:rl]{Step~4}).
Figure~\ref{fig:overview} summarizes the full pipeline.
Table~\ref{tab:notation} summarizes the principal notation used below.

\subsection*{Step 1: Training the Codec}
\phantomsection
\addcontentsline{toc}{subsection}{Step 1: Training the Codec}
\label{sec:method:latent}

\subsubsection{Action space.}
We work with a bimanual platform whose two $7$-DoF arms carry two $20$-DoF dexterous hands.
We represent a single-frame joint-space action by $a \in \R^{54}$, ordered as
\begin{equation}
\begin{aligned}
  a = \bigl(&a^{\mathrm{arm}}_{\mathrm{L}},\;
              a^{\mathrm{hand}}_{\mathrm{L}},\\
            &a^{\mathrm{arm}}_{\mathrm{R}},\;
              a^{\mathrm{hand}}_{\mathrm{R}}\bigr).
\end{aligned}
\label{eq:action-layout}
\end{equation}
The four blocks have dimensions $7$, $20$, $7$ and $20$, respectively.
Action chunks preserve this ordering at every time step, and the measured proprioceptive state $q \in \R^{54}$ follows the same ordering.

\subsubsection{Chunk-level hand action encoding.}
Motivated by the low-dimensional postural synergies observed in human grasps \citep{santello1998postural}, we compress each hand-action chunk from $\R^{H \times 20}$ to a latent sequence in $\R^{H \times 9}$ using a variational autoencoder \citep{kingma2014vae} trained separately for each hand and frozen thereafter.
We describe one hand below and suppress the side label.
Let $q_0 \in \R^{54}$ be the measured joint configuration at the start of the chunk, and let $q^{\mathrm{hand}}_0 \in \R^{20}$ denote the corresponding hand block.
Let $q^{\mathrm{hand,cmd}}_{1:H} \in \R^{H \times 20}$ denote the commanded joint-position chunk for that hand.
We represent it relative to the initial hand configuration:
\begin{equation}
  a^{\mathrm{hand}}_t = q^{\mathrm{hand,cmd}}_t - q^{\mathrm{hand}}_0,
  \qquad t = 1, \ldots, H.
  \label{eq:relative-hand-command}
\end{equation}
Each joint coordinate of $a^{\mathrm{hand}}_{1:H}$ is standardized before encoding; we omit this preprocessing from the notation.
Encoder and decoder are small temporal Transformers that attend across all $H$ frames of a chunk; their architecture is summarized in Table~\ref{tab:model_zoo} of Appendix~\ref{app:models}.
They use temporal embeddings fixed to $H$, and we apply the codec only to full-horizon chunks.
The encoder emits a per-frame Gaussian posterior, and the decoder reconstructs the hand action:
\begin{equation}
\begin{aligned}
  (\mu, \sigma) &= E^{\mathrm{hand}}(a^{\mathrm{hand}}_{1:H},\,
                                      q^{\mathrm{hand}}_0), \\
  z^{\mathrm{hand}}_{1:H} &= \mu + \sigma \odot \epsilon, \quad
    \epsilon \sim \mathcal{N}(0, I), \\
  \widehat{a}^{\mathrm{hand}}_{1:H}
    &= D^{\mathrm{hand}}(z^{\mathrm{hand}}_{1:H},\, q^{\mathrm{hand}}_0),
\end{aligned}
\label{eq:vae}
\end{equation}
with $\mu, \sigma, z^{\mathrm{hand}}_{1:H} \in \R^{H \times 9}$.
The codec is trained with
\begin{equation}
\begin{aligned}
  \mathcal{L}_{\mathrm{codec}}={}&\sum_{k=0}^{2}\lambda_k\mathcal{L}_{\mathrm{WH}}^{(k)} \\
  &+\beta_{\mathrm{KL}}\,\mathrm{KL}\!\left(\mathcal{N}(\mu,\sigma^2)\,\|\,\mathcal{N}(0,I)\right),
\end{aligned}
  \label{eq:codec-loss}
\end{equation}
Here $\Delta^0x_t=x_t$ and, for $k\in\{1,2\}$,
$\Delta^kx_t=\Delta^{k-1}x_{t+1}-\Delta^{k-1}x_t$; thus
$\Delta^1x_t=x_{t+1}-x_t$ and
$\Delta^2x_t=x_{t+2}-2x_{t+1}+x_t$.
All differences and errors are computed in the standardized relative-action space.
For temporal order $k\in\{0,1,2\}$, define the error at each valid frame $t\in\{1,\ldots,H-k\}$ and joint $j$ as
\begin{equation}
  e^{(k)}_{t,j}
  = \bigl(\Delta^k\widehat{a}^{\mathrm{hand}}\bigr)_{t,j}
    - \bigl(\Delta^ka^{\mathrm{hand}}\bigr)_{t,j}.
  \label{eq:codec-error}
\end{equation}
The per-joint weighted Huber term is
\begin{equation}
  \mathcal{L}_{\mathrm{WH}}^{(k)}
  = \frac{1}{20(H-k)}
    \sum_{t=1}^{H-k}\sum_{j=1}^{20}w_j\ell_{\tau_{\mathrm{Huber}}}\!\left(e^{(k)}_{t,j}\right),
  \label{eq:weighted-huber}
\end{equation}
where $w_j$ is the weight of joint $j$ and the scalar Huber penalty \citep{huber1964robust} is
\begin{multline}
  \ell_{\tau_{\mathrm{Huber}}}(e)=\\
  \begin{cases}
    \tfrac{1}{2}e^2, & |e|\leq\tau_{\mathrm{Huber}},\\
    \tau_{\mathrm{Huber}}\bigl(|e|-\tfrac{1}{2}\tau_{\mathrm{Huber}}\bigr), & |e|>\tau_{\mathrm{Huber}}.
  \end{cases}
  \label{eq:huber}
\end{multline}
Its quadratic branch remains sensitive to typical reconstruction errors, while its linear branch prevents occasional large deviations in the heterogeneous training corpus from dominating the update.
The $k=0$ term preserves joint positions, and the $k=1,2$ terms additionally preserve frame-to-frame motion and changes in that motion.
The threshold $\tau_{\mathrm{Huber}}$, joint weights $w_j$, and temporal weights $\lambda_k$ are specified in Appendix~\ref{app:schedules}.
The KL term regularizes the posterior toward $\mathcal{N}(0, I)$, fixing the latent scale and discouraging vanishing posterior variance; both are useful under the latent perturbations introduced in \hyperref[sec:method:rl]{Step~4}.
At inference, we use the posterior mean, $z^{\mathrm{hand}}_{1:H} = \mu$.

Combining the two hand codecs with the arm blocks gives the $54 \leftrightarrow 32$ codec $(E, D)$.
For a full action chunk $a_{1:H} \in \R^{H \times 54}$, the encoded chunk
$z = E(a_{1:H}; q_0) \in \R^{H \times 32}$ has the block structure
\begin{equation}
  z = \bigl(z^{\mathrm{arm}}_{\mathrm{L}},\;
            z^{\mathrm{hand}}_{\mathrm{L}},\;
            z^{\mathrm{arm}}_{\mathrm{R}},\;
            z^{\mathrm{hand}}_{\mathrm{R}}\bigr).
\label{eq:latent-layout}
\end{equation}
Each arm block is the corresponding $H \times 7$ action block, while each $H \times 9$ hand block is produced by \eqref{eq:vae}; thus the codec constrains only the hand degrees of freedom.

The decoder produces the uncompressed joint-space action $a_{1:H} = D(z; q_0)$.
Adding the starting state to each frame forms the executable command,
\[
q^{\mathrm{cmd}}_t = q_0 + a_t, \qquad t = 1, \ldots, H.
\]
Thus $a$ denotes the underlying joint-space action, whereas the learned policy operates in its encoded representation $z$.

\subsection*{Step 2: SFT}
\phantomsection
\addcontentsline{toc}{subsection}{Step 2: SFT}
\label{sec:method:sft}

\subsubsection{Policy in the latent action space.}
We fine-tune the open-source $\pi_{0.5}$ flow-matching vision-language-action model \citep{intelligence2025pi05}.
It maps an observation
\begin{equation}
  o = \bigl(I_{1{:}K},\; \ell,\; q\bigr),
  \label{eq:observation}
\end{equation}
to an action chunk of horizon $H = 32$, where $I_{1{:}K}$ are RGB images from $K = 3$ cameras, $\ell$ is the language instruction, and $q$ is the proprioceptive state
\citep{zhao2023act}. The observation is
processed by a pretrained vision-language backbone, while a noisy action chunk is carried by a smaller \emph{action expert} that attends to the backbone tokens and is trained by conditional flow matching \citep{lipman2023flow}.
A chunk is generated by integrating the predicted vector field from noise over a small number of Euler steps.
Inference is synchronous: of each chunk only the first $C = 25$ frames are executed before the next observation is taken, which trades the temporal consistency of open-loop chunk execution against reactivity
\citep{liu2025bid, zhao2026dehp}.

Rather than the $54$-dimensional joint-space action, the policy predicts the 32-dimensional representation of Eq.~\plaineqref{eq:latent-layout}, yielding $z = \pi(o) \in \R^{H \times 32}$.
At deployment the decoder reconstructs $a = D(z; q)$, and the robot executes $q + a_t$ at each frame of the chunk.
The encoder $E$ is used only during training to convert joint-space demonstrations into latent supervision targets.

\subsubsection{Supervising in both spaces.}
Because the hand decoder is nonlinear, equal errors in latent space can produce different joint-space errors.
We therefore augment the flow-matching loss with supervision on the decoded hand action.
Let $z$ be the latent target, $\epsilon \sim \mathcal{N}(0, I)$, and $t_{\mathrm{flow}}$ the flow time.
We construct $x_{t_{\mathrm{flow}}} = t_{\mathrm{flow}} \epsilon + (1 - t_{\mathrm{flow}}) z$ and train the action expert $v_\omega(x_{t_{\mathrm{flow}}}, t_{\mathrm{flow}}, o)$ against $u = \epsilon - z$.
Its one-step estimate of the clean chunk is $\widehat{z} = x_{t_{\mathrm{flow}}} - t_{\mathrm{flow}}\,v_\omega$, whose hand blocks decode to $\widehat{a}^{\mathrm{hand}}_{1:H} = D^{\mathrm{hand}}(\widehat{z}^{\,\mathrm{hand}}, q^{\mathrm{hand}}_0)$.
The objective is
\begin{equation}
\begin{aligned}
  \mathcal{L}_{\mathrm{sft}} =\;
    &\E \bigl[\, \lambda_{\mathrm{arm}} \lVert (v_\omega - u)^{\mathrm{arm}}
       \rVert^2 \\
    &\qquad\quad + \lambda_{\mathrm{lat}} \lVert (v_\omega - u)^{\mathrm{hand}}
       \rVert^2 \,\bigr] \\
    &+ \lambda_{\mathrm{dec}}\, \E
       \bigl\lVert \widehat{a}^{\mathrm{hand}}_{1:H}
                      - a^{\mathrm{hand}}_{1:H}
       \bigr\rVert_2^2 ,
\end{aligned}
\label{eq:sft-loss}
\end{equation}
Here, $\lambda_{\mathrm{arm}}$ weights flow matching on the $14$ uncompressed arm-joint coordinates, $\lambda_{\mathrm{lat}}$ on the $18$ hand-latent coordinates, and $\lambda_{\mathrm{dec}}$ on the decoded hand-action error.
Gradients from the decoded-action loss are backpropagated through the frozen decoder $D^{\mathrm{hand}}$ to update $\omega$.

\subsection*{Step 3: DAgger}
\phantomsection
\addcontentsline{toc}{subsection}{Step 3: DAgger}
\label{sec:method:dagger}
Supervised fine-tuning visits mostly expert-induced states, whereas in closed-loop execution small errors in approach, finger contact, or object pose alter subsequent observations and drive the robot out of that distribution.
We therefore aggregate operator corrections using human-gated DAgger \citep{ross2011dagger, kelly2019hgdagger}.

\begin{figure}[t]
\centering
\includegraphics[width=\columnwidth]{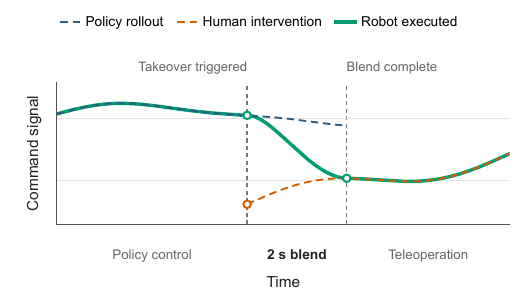}
\caption{Smooth hand-command transition during a DAgger takeover. Dashed curves show the policy and teleoperation commands, while the solid curve shows the command executed by the robot. At takeover, the executed command transitions over $2$\,s from the last policy command to the live teleoperation command and remains continuous across the switch.}
\label{fig:dagger-takeover}
\end{figure}

First, the last $5$\,s of robot state is held in a rolling buffer: when the operator judges that continued autonomous execution would fail or breach a safety boundary, they select a point inside that window and the robot is returned to it, so the correction starts before the error compounded rather than at the moment it became visible.
Second, an interface displays the operator's pose alongside the robot's to help the operator align the two before takeover.
Once takeover is triggered, we smoothly blend the hand-joint command over $2$\,s from the last command sent to the robot to the live teleoperation command, preventing a step change at the control switch.
Let $q^{\mathrm{hold}}$ denote the last hand-joint command sent before takeover and $q^{\mathrm{teleop}}_t$ the live operator command.  For elapsed blend time $s$ and blend duration $\tau_b=2$\,s, the hand command sent to the robot is the linear interpolation
\begin{equation}
  \begin{aligned}
    q^{\mathrm{hand,cmd}}_t
      &= (1-\alpha_s)q^{\mathrm{hold}} + \alpha_s q^{\mathrm{teleop}}_t,\\
    \alpha_s &= \min\!\left\{\frac{s}{\tau_b},1\right\}.
  \end{aligned}
  \label{eq:dagger-blend}
\end{equation}
so that $\alpha_s=0$ at the start of takeover and $\alpha_s=1$ after the blend completes.
Figure~\ref{fig:dagger-takeover} illustrates this transition.
The operator then completes the episode.

For supervision, the live operator command is recorded throughout the blend rather than the blended command sent to the robot.
The retained joint-space trajectory is converted to the representation of \eqref{eq:action-layout} and, for the latent-action policy, encoded with $E$.
After each collection round, the resulting corrections are appended to the dataset, and the policy is retrained for 30\,k update steps from the same public $\pi_{0.5}$ checkpoint.
Each minibatch draws a fraction $\alpha_{\mathrm{new}}$ from the newly collected corrections and the remaining $1-\alpha_{\mathrm{new}}$ from the previously accumulated demonstrations and corrections.

\subsection*{Step 4: Real-world RL}
\phantomsection
\addcontentsline{toc}{subsection}{Step 4: Real-world RL}
\label{sec:method:rl}

\subsubsection{The RL token.}
To represent the observation, we use the RL token of \citet{xu2026rltoken}, a single vector $h_{\mathrm{rl}}$ summarizing what the frozen backbone has computed about $o$.
To obtain $h_{\mathrm{rl}}$, we take the final-layer image-token features $\eta_{1{:}M}$ from a prefix-only pass through the frozen backbone, append a learnable readout token $\xi_{\mathrm{rl}}$, and process the resulting sequence with a two-layer bidirectional transformer $g_\varphi$.
Each encoder block contains bidirectional self-attention, a second attention sublayer over the sequence entering that block, and a feed-forward sublayer.
The output at the appended readout position $M+1$ is then projected by a learned linear layer $W_{\mathrm{rl}}$ to obtain
\begin{equation}
  h_{\mathrm{rl}} = W_{\mathrm{rl}}\,
    g_\varphi\bigl([\,\eta_{1{:}M};\; \xi_{\mathrm{rl}}\,]\bigr)_{M+1}
    \in \R^{2048}.
  \label{eq:rltoken}
\end{equation}
The image-token features already incorporate the language instruction and proprioceptive state through the backbone's bidirectional prefix attention.
To retain this information in $h_{\mathrm{rl}}$, we train the readout on the reference policy's demonstration set $\mathcal{D}$ by reconstructing its image-token features with an auxiliary autoregressive decoder $d_\psi$, while keeping the backbone frozen.
The decoder uses teacher forcing with right-shifted target features, causal self-attention, and cross-attention to $h_{\mathrm{rl}}$.
Over the unpadded image-token positions $\mathcal{V}$, we average the squared reconstruction error across positions and their 2048 feature coordinates:
\begin{multline}
  \mathcal{L}_{\mathrm{ro}}(\varphi, \psi) = \\
  \E_{o \sim \mathcal{D}} \biggl[
    \frac{1}{2048\,\lvert \mathcal{V} \rvert}
    \sum_{i \in \mathcal{V}}
    \bigl\lVert d_\psi(h_{\mathrm{rl}},\, \eta_{<i}) - \eta_i \bigr\rVert_2^2
  \biggr],
  \label{eq:readout-loss}
\end{multline}
The prefix features are stop-gradiented, so only the readout token, encoder, projection, and auxiliary decoder receive gradient updates.
After training, we discard $d_\psi$ and freeze the readout parameters $(\xi_{\mathrm{rl}}, g_\varphi, W_{\mathrm{rl}})$.
During online RL, the frozen readout reuses the reference policy's prefix features to compute $h_{\mathrm{rl}}$ once per chunk.
Implementation details and the default offline pretraining configuration are given in Appendix~\ref{app:rlt-pretraining} and Table~\ref{tab:rlt-schedule}.

\subsubsection{The latent residual.}
\label{sec:method:residual}
Real-robot data is too scarce to continue training the resulting policy, and updating its parameters would eliminate it as an exact fallback.
We therefore freeze the resulting policy, denoted by $\pi_{\mathrm{ref}}$, and learn a bounded residual $\Delta z$ in the same space as its latent output $z_{\mathrm{ref}} = \pi_{\mathrm{ref}}(o)$.
The corrected latent action $z_{\mathrm{ref}} + \Delta z$ is passed through the frozen decoder $D$, so setting $\Delta z = 0$ exactly recovers the reference policy's original command.

\begin{samepage}
Let $f_\theta$ denote the MLP residual actor conditioned on $h_{\mathrm{rl}}$, $q$, and $z_{\mathrm{ref}}$.
The actor produces a bounded per-frame correction $c$, from which we form the full-horizon residual $\Delta z$ and the resulting joint-command chunk $q^{\mathrm{cmd}}_{1:H}$:
\begin{align}
  c_{i,d} &= \begin{cases}
       \begin{aligned}
       &b^{c}_{d} \tanh\!\left(
         \bigl[f_\theta(h_{\mathrm{rl}},q,z_{\mathrm{ref}})\bigr]_{i,d}
       \right), \\[-2pt]
       &\hspace{2em} 1 \leq i \leq C,
       \end{aligned}
         \\[6pt]
       0, \qquad C < i \leq H,
     \end{cases} \label{eq:actor} \\
  \Delta z_{i,d} &= \begin{cases}
       \displaystyle\sum_{j=1}^{i} c_{j,d},
         & \text{arm dimension }d, \\[4pt]
       c_{i,d},
         & \text{hand-latent dimension }d
     \end{cases} \label{eq:accumulate} \\
  q^{\mathrm{cmd}}_{1:H}
    &= q + D\bigl(z_{\mathrm{ref}} + \Delta z; q\bigr).
       \label{eq:main}
\end{align}
\end{samepage}
Here, $q$ is broadcast across the $H$ frames.

The bound $b^{c}_{d}$ sets the allowed per-frame correction for each action-dimension group, with $b^{c}_{d} = b^{c}_{\mathrm{arm}}$ for the arm dimensions and $b^{c}_{d} = b^{c}_{\mathrm{hand}}$ for the hand-latent dimensions.
The actor's output layer is zero-initialised, so training starts at $\Delta z = 0$.
We apply a task-specific residual mask before adding the correction to the reference action.

\begin{equation}
  \Delta z \leftarrow m_{\mathrm{task}} \odot \Delta z .
  \label{eq:task-residual-mask}
\end{equation}

Here, $\odot$ denotes element-wise multiplication, and $m_{\mathrm{task}}$ is a binary mask over the residual action dimensions, broadcast across the $H$ frames.

For each hand, the residual modifies nine latent coordinates per active frame, and the full latent sequence is decoded jointly into joint commands.
This provides a compact, learned parameterisation of hand-motion corrections.

\subsubsection{Chunk-level residual TD3.}
Each RL transition spans the $C$ frames executed between two policy queries.
At chunk $t$, the actor takes $h_{\mathrm{rl},t}$, $q_t$, and $z_{\mathrm{ref},t}$ as input and produces $c_{1:C}$, from which the full-horizon latent residual action $\Delta z_t$ is formed as defined above.
The resulting latent chunk $z_{\mathrm{ref},t} + \Delta z_t$ is decoded and executed for $C$ frames.
After each episode, the operator labels the outcome; the reward assigned to its terminal chunk is $r_T=-1$ for failure and $r_T=0$ for success, while all preceding chunks receive zero reward.

We optimise the residual with TD3 \citep{fujimoto2018td3}, whose deterministic actor and twin critics operate at this chunk-level time scale.
We maintain target copies of the actor and twin critics, denoted $f_{\bar{\theta}}$ and $Q_{\bar{\phi}_1}, Q_{\bar{\phi}_2}$, and update them by Polyak averaging.
The twin critics $Q_{\phi_1}$ and $Q_{\phi_2}$ take $(h_{\mathrm{rl}}, q, z_{\mathrm{ref}} + \Delta z)$ as input and regress on the clipped double-Q target
\begin{equation}
\begin{aligned}
  y_t ={}& r_t + (1 - d_t)\, \gamma^{\,C}
      \min_{i \in \{1, 2\}} \\
      &Q_{\bar{\phi}_i}\bigl(h_{\mathrm{rl}}',\, q',\,
      z_{\mathrm{ref}}' + \Delta z'\bigr),
\end{aligned}
  \label{eq:critic-target}
\end{equation}
Here, $d_t=1$ for the terminal transition and $d_t=0$ otherwise.
The per-frame discount $\gamma$ gives a discount of $\gamma^{\,C}$ over each $C$-frame transition.
Primes denote quantities at the next chunk, and $\Delta z'$ is formed using the target actor $f_{\bar{\theta}}$.

During collection, we add exploration noise to the bounded correction $c_{1:C}$, yielding $\tilde{c}_i = c_i + \sigma \varepsilon_i$, and form $\Delta z$ from $\tilde{c}_{1:C}$.
We use separate noise scales $\sigma^{\mathrm{arm}}$ and $\sigma^{\mathrm{hand}}$ for the arm and hand-latent dimensions, respectively.
For exploration within a chunk, we add white noise to the arm increments and accumulate these increments across the chunk, whereas the hand latents receive per-dimension normalised Brownian noise directly, with spectral exponent $\nu=2$ and a $1/f^2$ power spectrum \citep{eberhard2023pink}.
Before execution, the perturbed chunk passes through the downstream safety clamps; evaluation uses no exploration noise.

A replay transition stores $(h_{\mathrm{rl}}, q, z_{\mathrm{ref}}, \Delta z, r, d)$, together with the corresponding next-chunk quantities $(h_{\mathrm{rl}}', q', z_{\mathrm{ref}}')$.
Here, the stored $\Delta z$ is the task-masked residual formed during collection, including exploration noise.
For critic updates, the sampled latent action is reconstructed as $z_{\mathrm{ref}} + \Delta z$.
We define the residual-magnitude regulariser on the task-masked, full-horizon latent residual as
\begin{equation}
  \mathcal{R}_{\Delta z}(\Delta z)
    \mathrel{:=} \lVert\Delta z\rVert_{F}^{2}
    = \sum_{i=1}^{H}\sum_{d=1}^{32}(\Delta z_{i,d})^{2}.
  \label{eq:residual-regulariser}
\end{equation}
The norm covers all $H$ frames, with inactive dimensions contributing nothing.
Beyond $C$, hand-latent residuals are zero, while active arm residuals retain their final accumulated values and remain included in the regulariser.
The critics regress on \eqref{eq:critic-target} at every step; once joint training is enabled, the actor is updated after every critic update, minimising
\begin{align}
  &- \beta_{q}\, Q_{\phi_1}\!\left(h_{\mathrm{rl}}, q,
  z_{\mathrm{ref}} + \Delta z\right)
  \notag \\
  &\quad + \beta_{\Delta z}\, \mathcal{R}_{\Delta z}(\Delta z)
  \notag \\
  &\quad + \beta_{s}\, \mathcal{S}(z_{\mathrm{ref}} + \Delta z),
  \label{eq:actor-loss}
\end{align}
For a latent chunk $z_{1:H}$, the temporal smoothness penalty is
\begin{equation}
  \mathcal{S}(z_{1:H})
  = \frac{1}{H-2}\sum_{i=1}^{H-2}
    \left\lVert z_{i+2} - 2z_{i+1} + z_i \right\rVert_2^2.
  \label{eq:temporal-smoothness}
\end{equation}
In \eqref{eq:actor-loss}, $\Delta z$ is recomputed without exploration noise using the current actor and the stored $(h_{\mathrm{rl}},q,z_{\mathrm{ref}})$, and $z_{1:H}=z_{\mathrm{ref}}+\Delta z$.
The three terms respectively maximise the critic value, penalise the residual magnitude through $\mathcal{R}_{\Delta z}$, and penalise the second temporal difference of this resulting latent chunk.
The actor objective is evaluated entirely in the latent space, so its gradients do not pass through $D$.
The loss weights and remaining RL hyperparameters are listed in Table~\ref{tab:rl-schedule} of Appendix~\ref{app:schedules}.

\subsubsection{Online post-training protocol.}
Online post-training proceeds in two phases.
During the \emph{warm-up} phase, the actor output is set to zero, exploration noise is applied, and only the critics train while the operator watches their value estimates on live rollouts.
The operator manually starts the \emph{joint training} phase with a keystroke, after confirming that critic training has stabilised and the value estimates distinguish success from failure.
In this phase, the actor output is executed with exploration noise.
The critics continue to train, and the actor is updated after every critic update.

Before decoding, we form the full $H$-frame residual $\Delta z$ from the actor's $C$-frame output as in \eqref{eq:accumulate} and add it to $z_{\mathrm{ref}}$.
Three safety bounds are applied before execution.
First, $b^{c}$ bounds the per-frame correction $c$ before exploration.
Second, the hand latents in $z_{\mathrm{ref}} + \Delta z$ are clipped to the 99.5th-percentile envelope $b^{z}$ recorded during VAE training.
Third, the final joint-position command is rate-limited to a maximum change of
$\dot q_{\max}$ per control frame.

\begin{algorithm*}[t]
\caption{Online post-training with critic warm-up followed by joint actor--critic training.}
\label{alg:dexlrl}
\begin{algorithmic}[1]
\Require frozen $\pi_{\mathrm{ref}}$, decoder $D$, RL-token encoder $g_\varphi$
\Require actor $f_\theta$, critics $Q_{\phi_1}, Q_{\phi_2}$, buffer
         $\mathcal{B}$, phase $P \in \{\textsc{warmup}, \textsc{joint}\}$
         initialised to \textsc{warmup}
\Statex
\Statex \textbf{Serving loop} \Comment{one iteration per action chunk}
\For{each episode}
  \For{$t = 1, 2, \dots$ until the operator ends the episode}
    \State observe $o_t = (I_{1:K}, \ell, q_t)$
    \State $h_{\mathrm{rl}} \gets$ readout of $g_\varphi$ on the prefix of
           $\pi_{\mathrm{ref}}(o_t)$ \Comment{\eqref{eq:rltoken}}
    \State $z_{\mathrm{ref}} \gets \pi_{\mathrm{ref}}(o_t) \in \R^{H \times 32}$
           \Comment{no encoding}
    \If{$P = \textsc{joint}$}
      \State $c \gets b^{c} \cdot \tanh\bigl(f_\theta(h_{\mathrm{rl}}, q_t,
             z_{\mathrm{ref}})\bigr)$ \Comment{\eqref{eq:actor}}
    \Else
      \State $c \gets 0$ \Comment{actor output disabled}
    \EndIf
    \State $c \gets c + \sigma \epsilon$
           \Comment{white arm noise on increments; direct Brownian $\nu{=}2$ hand noise}
    \State form task-masked $\Delta z$ from $c$ using \eqref{eq:accumulate} and \eqref{eq:task-residual-mask}
    \State $z \gets z_{\mathrm{ref}} + \Delta z$; clamp
           $z^{\mathrm{hand}}$ to $b^{z}$
    \State $q^{\mathrm{cmd}}_{1:H}
           \gets q_t + D(z; q_t)$
           \Comment{$q_t$ broadcast over $H$ frames}
    \State rate-limit $q^{\mathrm{cmd}}_{1:H}$ to a maximum change of
           $\dot q_{\max}$ per control frame
    \State execute the first $C$ frames; record the transition
           \Comment{store $z_{\mathrm{ref}}$ and $\Delta z$ separately}
  \EndFor
  \State operator labels the episode outcome
  \State assign $r=-1$ to the terminal transition on failure and $r=0$
         to all other transitions; append them to $\mathcal{B}$
\EndFor
\Statex
\Statex \textbf{Learner} \Comment{runs once $|\mathcal{B}|$ exceeds a minimum size}
\While{post-training is running}
  \State sample a batch from $\mathcal{B}$
  \State update $\phi_1, \phi_2$ on the target $y$ of
         \eqref{eq:critic-target} \Comment{critic action: stored $z_{\mathrm{ref}}+\Delta z$}
  \If{$P = \textsc{joint}$}
    \State recompute $\Delta z$ using the current actor without exploration noise
    \State after each critic update, update $\theta$ on
           \eqref{eq:actor-loss} and export a snapshot
  \EndIf
  \State Polyak-update the target critic; if $P=\textsc{joint}$, also update the target actor
\EndWhile
\end{algorithmic}
\end{algorithm*}

\section{Experiments}
\label{sec:exp}

We evaluate the proposed post-training pipeline and its individual stages on five real-world dexterous manipulation tasks, testing each reported checkpoint over 20 trials.
The final policy for every task succeeds in all 20 evaluation trials.

\subsection{Experimental setup}
\label{sec:exp:config}

\subsubsection{Hardware and control.}
\begin{figure}[t]
\centering
\includegraphics[width=\columnwidth]{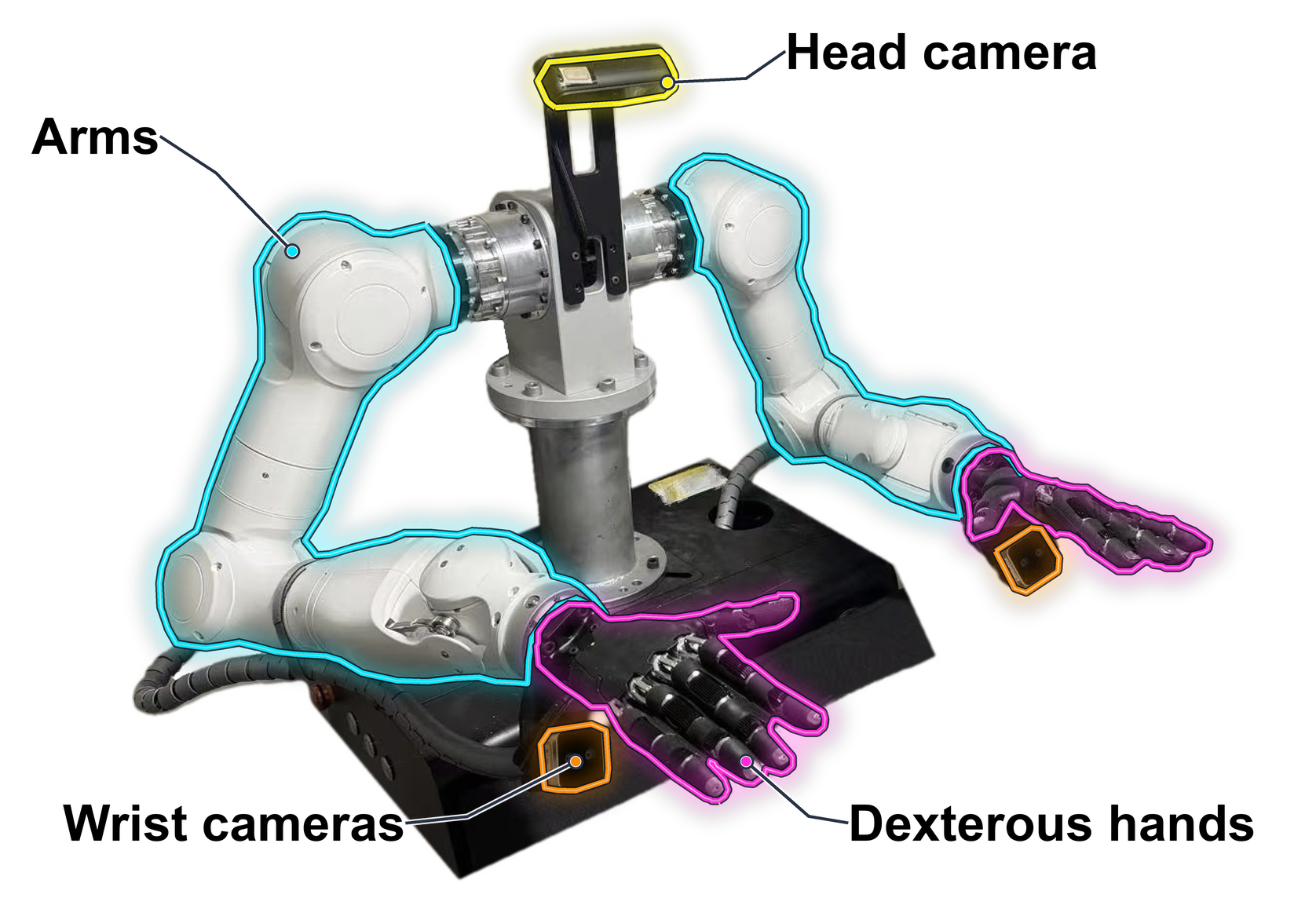}
\caption{Robot platform used for data collection and evaluation.
The system comprises two 7-DoF Tianji arms, two 20-DoF Wuji dexterous hands, one RGB wrist camera mounted on each arm, and one RGB head camera.}
\label{fig:robot-platform}
\end{figure}

We use a bimanual platform with two 7-DoF arms from Tianji and two 20-DoF hands from Wuji (Figure~\ref{fig:robot-platform}).
We use one head-mounted RGB camera and two wrist-mounted RGB cameras, with one wrist camera attached to each arm.
We teleoperate the arms with a Vive Tracker and the hands using both Wuji Glove and Manus gloves.
Both the arms and hands are controlled in joint position at 30\,Hz.
We record the three camera views, joint commands, and measured joint states at 30\,Hz as well.
The final joint-position command is rate-limited to a maximum change of
$\pi/18\,\mathrm{rad}$ per control frame.

\subsubsection{Tasks.}
\begin{figure*}[t]
\centering
\includegraphics[width=\textwidth]{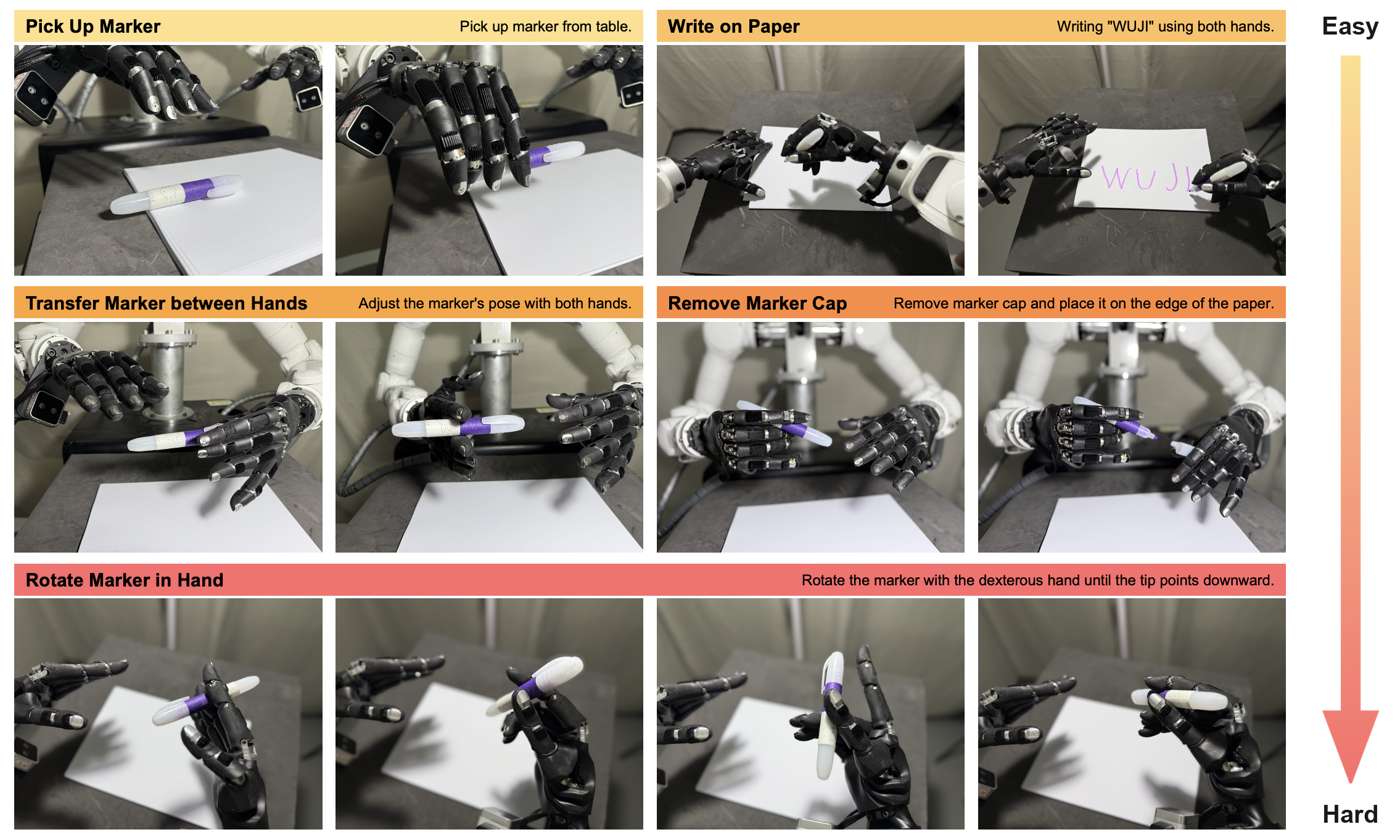}
\caption{The five marker-manipulation tasks used for evaluation: Pick Up Marker, Transfer Marker between Hands, Rotate Marker in Hand, Remove Marker Cap, and Write on Paper.
The tasks cover grasping, bimanual transfer, in-hand manipulation, object separation, and contact-rich writing.
The yellow-to-red arrow orders the tasks from easier to harder.}
\label{fig:evaluation-tasks}
\end{figure*}

The five tasks illustrated in Figure~\ref{fig:evaluation-tasks} use the same marker and span grasping, bimanual transfer, in-hand manipulation, object separation, and contact-rich writing.
Each task uses an operator-checked success criterion.
Videos of these tasks are available on the \href{https://wuji.tech/blog/post-training-0}{project website}.

\begin{itemize}
  \item \emph{Pick Up Marker} \mbox{(\emph{Pick up})} succeeds when the robot lifts the marker completely off the table and ends with the marker securely grasped.
  \item \emph{Transfer Marker between Hands} \mbox{(\emph{Transfer})} succeeds when the marker is passed from the right hand into a stable grasp in the left hand and is then secured between the right index and middle fingers.
  \item \emph{Rotate Marker in Hand} \mbox{(\emph{Rotate})} succeeds when the right hand rotates the marker without dropping it and finishes with its tip pointing downward.
  \item \emph{Remove Marker Cap} \mbox{(\emph{Uncap})} succeeds when the left hand completely separates the cap from the marker without dropping either component.
  \item \emph{Write on Paper} \mbox{(\emph{Write})} succeeds when the robot writes the complete word ``WUJI,'' the brand name of the dexterous hands used in our setup, while maintaining contact between the marker tip and the paper.
\end{itemize}

Across these tasks, task-specific adaptation begins with supervised fine-tuning.
Pick Up Marker, Remove Marker Cap, and Write on Paper reach 100\% success after DAgger.
Transfer Marker between Hands requires DAgger followed by latent residual RL, whereas Rotate Marker in Hand proceeds directly from supervised fine-tuning to latent residual RL because its physical object state cannot be restored during takeover.

Every trial is reset manually.
Pick Up Marker starts with the marker lying on the right half of the table.
Transfer Marker between Hands starts with the marker grasped in the right hand.
Rotate Marker in Hand starts with the marker held between the right index and middle fingers.
Remove Marker Cap starts with the capped marker already held by the robot.
Write on Paper starts with the marker already held by the robot and a sheet of paper placed at the centre of the table.
Each checkpoint is evaluated over 20 task trials under the corresponding task-specific reset condition, and its success rate is the fraction of successful trials.
Training random-seed settings are held fixed across the dataset comparisons, while the training data vary with the collection stage and budget.

Lighting is held approximately constant throughout evaluation, while for each task, the initial object pose or grasp is manually varied within the range represented in the data-collection distribution.

\subsection{Training the hand action codec}
\label{sec:exp:codec}

\subsubsection{Training corpus.}
The codec is trained on three data sources.
First, our internal teleoperation corpus contains approximately 38M frames per hand, collected across robots in the same family and a range of manipulation tasks.
Second, we use the VITRA-1M human-hand trajectories reconstructed from egocentric video and retargeted to our 20-DoF hand \citep{li2026vitra}.
Third, we include the task-specific demonstration set used in \hyperref[sec:method:sft]{Step~2}, with 200 episodes per task.
The task-specific codec data are restricted to these initial 200 demonstrations per task and exclude all subsequent DAgger corrections and online RL data.
We train separate codecs for the two hands on the combined corpus.
Full optimisation and loss settings are provided in Appendix~\ref{app:schedules}.

\subsubsection{Comparison of compression methods.}
To assess how much of the hand's motion each representation can preserve, we compare the reconstruction quality of several compression methods, all evaluated at nine latent dimensions per hand.
The methods differ in linearity, temporal context, and architecture.
Principal component analysis (PCA) provides a frame-wise linear baseline motivated by postural synergies in hand configurations \citep{santello1998postural}.
It independently projects each standardised 20-dimensional hand action onto nine principal components and reconstructs it without modelling dependencies across frames.
Two frame-based autoencoders --- an MLP and a Transformer-based VAE --- encode each frame in isolation and therefore carry no temporal context.
The remaining two operate on whole chunks: a chunk-based MLP, which receives the chunk flattened into a single vector, and our chunk-level variational Transformer, which attends across the $H$ frames of a chunk and serves as the codec of \hyperref[sec:method:latent]{Step~1}.

All methods are trained on the same corpus and evaluated on the same held-out validation set.
We report the reconstruction error in degrees per joint, averaged over the 40 joints of both hands, in Figure~\ref{fig:codec-analysis}(a).

\subsubsection{Choosing the latent dimensionality.}
\begin{figure}[t]
\centering
\includegraphics[width=\columnwidth]{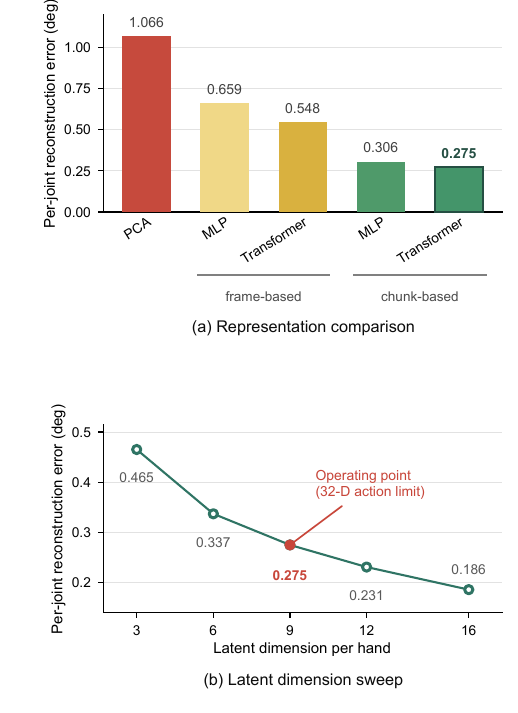}
\caption{Hand-action codec analysis.
Panel (a) compares the reconstruction error of five compression methods, all using nine latent dimensions per hand.
Panel (b) shows the reconstruction error of the chunk-level Transformer codec as the latent width varies.
All errors are measured on the same held-out validation set, reported in degrees per joint, and pooled across the 40 joints of both hands.
The nine-dimensional operating point is selected to satisfy the pretrained policy's 32-dimensional action interface.}
\label{fig:codec-analysis}
\end{figure}

With the architecture fixed, we sweep the latent width of the chunk-level Transformer and measure the reconstruction error on the same held-out validation set, reported in Figure~\ref{fig:codec-analysis}(b).
Two considerations set the operating point.
The first is accuracy, which improves monotonically with width but at a diminishing rate: the three dimensions from six to nine remove about 1.4 times as much error as the three from nine to twelve do, so each further dimension buys less reconstruction quality than the one before it.

The second is a constraint from the reference policy: at nine dimensions per hand the layout of \eqref{eq:latent-layout} totals $2\times7 + 2\times9 = 32$ dimensions, which is the largest action dimensionality the pretrained $\pi_{0.5}$ checkpoint accepts.
Nine dimensions per hand meets both, and is the width used in every experiment reported here.

\subsection{Supervised fine-tuning and data aggregation}
\label{sec:exp:sft}

We first supervised fine-tune separate policies on 50, 100, 150, 200, 250 and 300 demonstrations, each for 30\,k update steps.
Using the 200-demonstration policy for the initial collection, we then collect two rounds of 50 corrections using the takeover procedure of \hyperref[sec:method:dagger]{Step~3}, with $\alpha_{\mathrm{new}}=0.5$.
After each round, we reinitialise the VLA from the same public $\pi_{0.5}$ checkpoint and train for 30\,k update steps on the aggregated dataset.
All policies use a batch size of 64; Appendix~\ref{app:schedules} gives the remaining optimisation details.
We evaluate DAgger on the four tasks that admit takeover; Rotate Marker in Hand is evaluated under SFT only.

\begin{figure*}[t]
\centering
\includegraphics[width=\textwidth]{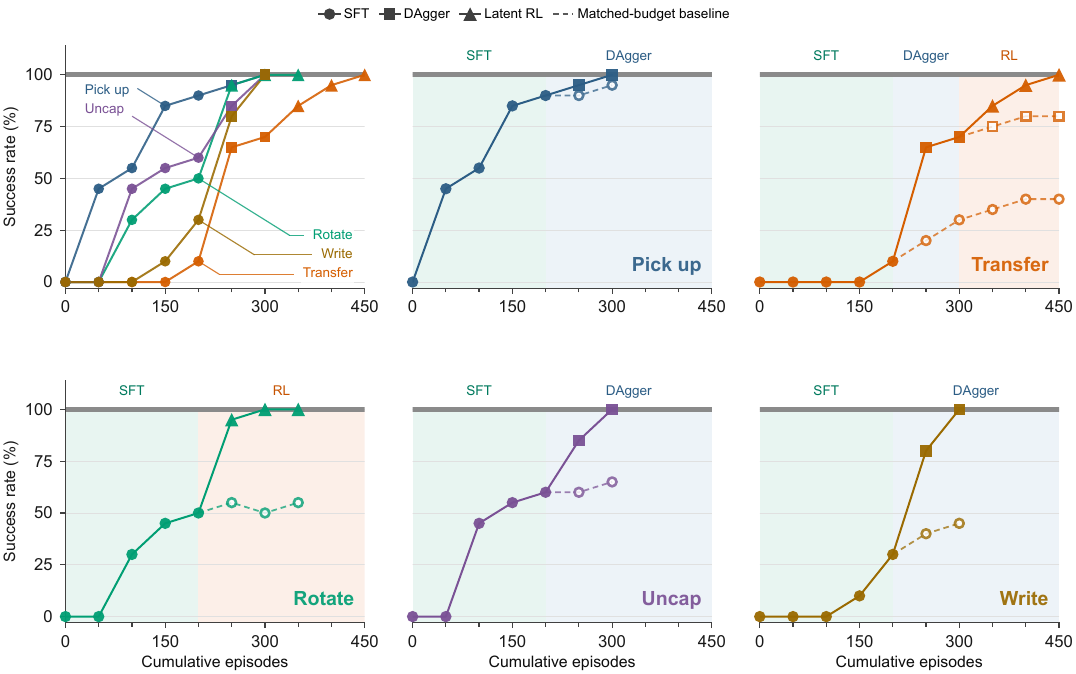}
\caption{Success rate as a function of cumulative episodes collected.
The overview panel overlays the trajectories for all five tasks, labelled directly on their curves; the remaining panels show the tasks individually.
Marker shapes identify training methods: circles for SFT, squares for DAgger, and triangles for latent residual RL.
Solid lines with filled markers show our staged pipeline; dashed lines with open markers show matched-budget baselines that continue the indicated method.
Background shading marks the stage reached by the main pipeline.
Each checkpoint is evaluated over 20 trials.
Rotate Marker in Hand admits no corrections and moves directly from SFT to online RL.}
\label{fig:sft}
\end{figure*}

For the continued-imitation controls in Figure~\ref{fig:sft}, we additionally train demonstration-only policies on 350, 400 and 450 demonstrations for Transfer Marker between Hands and on 350 demonstrations for Rotate Marker in Hand.
The continued-DAgger baseline for Transfer Marker between Hands adds three further rounds of 50 corrections after the 300-episode checkpoint, yielding 350, 400 and 450 cumulative episodes.
Each additional correction round uses the policy trained on the preceding aggregated dataset for collection.
All continued-imitation checkpoints follow the same 30\,k-update training schedule from the public $\pi_{0.5}$ checkpoint, with $\alpha_{\mathrm{new}}=0.5$ for DAgger rounds and no online RL.
The three analyses below build on this experimental setup.

\medskip
\noindent\textit{1) Are DAgger corrections more sample-efficient than additional demonstrations?}\par

Figure~\ref{fig:sft} compares each DAgger checkpoint with a demonstration-only policy trained from the same cumulative episode budget.
At both 250 and 300 episodes, the DAgger pipeline achieves higher success across the four tasks that admit takeover.

Additional demonstrations are collected independently of the current policy and therefore do not adapt to its changing failure modes.
DAgger instead uses rollouts of the current policy to identify where corrections are needed.
At each intervention, the operator provides a recovery trajectory from a buffered pre-failure state, so successive rounds focus on the errors that remain.

\medskip
\noindent\textit{2) Does imitation in the hand-latent space preserve raw joint-space performance?}\par
\begin{figure}[t]
\centering
\includegraphics[width=\columnwidth]{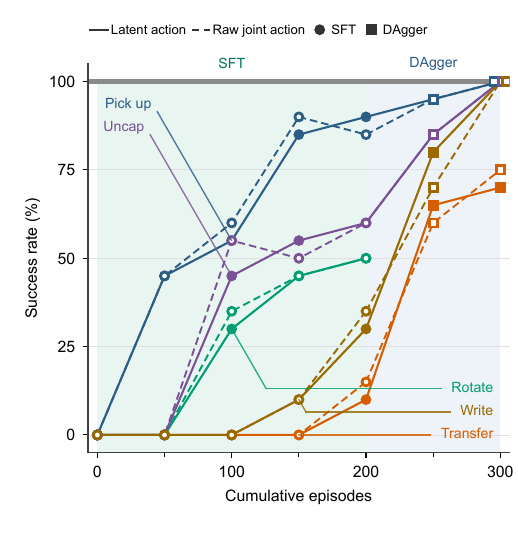}
\caption{Imitation learning in the raw joint space and the learned hand-latent space across the five tasks.
}
\label{fig:latent-imitation}
\end{figure}

To isolate the effect of the hand-latent representation, we train a raw-action baseline by expanding the action head from 32 to 54 dimensions, allowing it to predict joint-space actions directly without the codec.
Both variants use the same SFT demonstration subsets and a matched DAgger budget of two 50-correction rounds from their respective 200-demonstration checkpoints.
All other training and evaluation settings are held fixed.
Their success-rate curves remain closely aligned, showing no consistent loss in imitation performance from compressing each hand from 20 to 9 dimensions (Figure~\ref{fig:latent-imitation}).

\medskip
\noindent\textit{3) Where does imitation learning remain insufficient?}\par
Data aggregation does not close the gap on every task, and on Transfer Marker between Hands, Remove Marker Cap, and Write on Paper the gain from the second 50-episode correction round is smaller than from the first, indicating that improvement slows as aggregation proceeds; Pick Up Marker is already near saturation after SFT.
Transfer Marker between Hands rises from 10\% to 65\% after the first 50 corrections and to 70\% after the second round, but remains short of full success.
Rotate Marker in Hand reaches 50\% under SFT and cannot use DAgger because its physical object state cannot be restored at takeover.
These remaining gaps motivate the online residual RL evaluated in Section~\ref{sec:exp:rl}.

\subsection{Online residual reinforcement learning}
\label{sec:exp:rl}

Online post-training is evaluated on Transfer Marker between Hands and Rotate Marker in Hand, the two tasks left below full success after imitation learning.
Transfer Marker between Hands starts from the 70\% DAgger checkpoint, whereas Rotate Marker in Hand starts from the 50\% SFT checkpoint.
For Transfer Marker between Hands, the task-specific mask in \eqref{eq:task-residual-mask} selects the full 32-dimensional latent action, comprising both 7-DoF arms and both 9-dimensional hand latents.
For Rotate Marker in Hand, the mask selects only the 9-dimensional latent of the active right hand and fixes every arm and left-hand residual dimension to zero.
The two analyses below use these settings.

\medskip
\noindent\textit{1) How do latent and raw-space residual RL compare with the frozen policy?}\par

To isolate the effect of the residual action space, we compare the frozen policy, a raw-space residual policy \citep{johannink2019residual}, and the latent-residual policy of \hyperref[sec:method:rl]{Step~4}.
The raw-space actor adds its residual directly to the joint-position action chunk produced by the frozen VLA decoder, whereas the latent-space actor adds its residual to $z_{\mathrm{ref}}$ before the same frozen decoder.
Specifically, letting $a_{\mathrm{ref}}=D(z_{\mathrm{ref}};q)$ denote the decoded reference action, the raw-space actor takes $h_{\mathrm{rl}}\in\R^{2048}$, $q\in\R^{54}$, and $a_{\mathrm{ref},1:C}\in\R^{C\times54}$ as input and outputs a $C\times54$ joint-space residual:
\begin{equation}
\begin{aligned}
  \Delta^{\mathrm{raw}}_{1:C}
    &= b^{\mathrm{raw}} \odot \tanh\!\left(
       f_{\psi}^{\mathrm{raw}}(h_{\mathrm{rl}},q,a_{\mathrm{ref},1:C})
       \right) \\
    &\in \R^{C \times 54}, \\
  q^{\mathrm{cmd,raw}}_{1:C}
    &= q + a_{\mathrm{ref},1:C} + \Delta^{\mathrm{raw}}_{1:C}.
\end{aligned}
\label{eq:raw-residual}
\end{equation}
Thus, the raw actor outputs a bounded residual in the decoded joint-action space, with no hand-latent decoding in the residual path.
Both variants otherwise use the same TD3 training and execution protocol, online interaction budget, task-specific residual scope, and safety limits.
They also use the same temporal exploration-noise construction: white noise is added to the arm increments and accumulated across the chunk, while direct per-dimension Brownian noise with $\nu=2$ is applied to the hand block.
The noise amplitudes are chosen to match the magnitude of the resulting joint-space perturbations, measured after decoding for the latent variant (Appendix~\ref{app:schedules}).

\begin{figure}[t]
\centering
\includegraphics[width=\columnwidth]{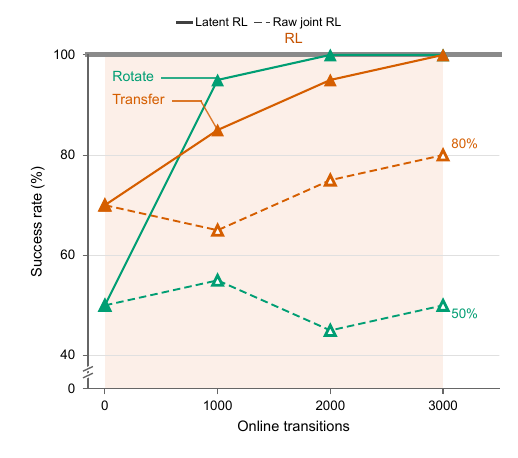}

\caption{Comparison of latent residual RL and raw-space residual RL over online interaction.}
\label{fig:rl-curves}
\end{figure}

Figure~\ref{fig:rl-curves} follows both residual methods over 3000 online transitions.
At the 30\,Hz control rate, each transition spans $C=25$ executed frames, or approximately $0.83$\,s of commanded motion.
The latent-residual policy improves Transfer Marker between Hands from 70\% to 100\% and Rotate Marker in Hand from 50\% to 100\%.
See the videos on the \href{https://wuji.tech/blog/post-training-0}{project website} for the post-trained VLA executing both tasks.
The raw-residual policy ends at 80\% and 50\%, respectively, and its success rate fluctuates rather than increasing consistently over training.
Qualitatively, under exploration, its joint-space perturbations often appear as uncoordinated motion across individual joints rather than a coherent corrective action.

The performance gap between the two residual policies can be understood by examining how their action spaces shape exploration.
The raw residual can perturb the active joint coordinates independently, whereas the latent residual is decoded from a hand representation learned from recorded motion.
The latent action space may therefore bias exploration toward coordinated hand motion and reduce the unstructured joint-space directions that must be explored online.
The contrast is clearest on Rotate Marker in Hand, where the correction acts entirely on the hand.

Across multiple evaluation trials, the latent-residual policy produces corrective behaviours not observed in our frozen-policy trials, including regrasping, maintaining contact, and recovering from incipient errors.
Their recurrence suggests that the improvement reflects learned corrective responses rather than isolated favourable perturbations.
Additional trials with randomized initial positions and external disturbances applied to the object and robot also provide evidence of robustness beyond the nominal evaluation condition.
Within these two tasks and the tested interaction budget, latent residual RL reaches full success, whereas raw-space residual RL does not.

\medskip
\noindent\textit{2) How efficiently does latent residual RL close the remaining imitation gap?}\par

\textbf{Sample efficiency.}
Figure~\ref{fig:sft} compares SFT, DAgger, and latent residual RL on a common cumulative-episode axis for Transfer Marker between Hands and Rotate Marker in Hand.
For the RL curves, 1000 online transitions correspond to 50 rollout episodes.
The latent-residual policy surpasses the frozen policy at that first evaluation and reaches 100\% after 100 episodes on Rotate Marker in Hand and 150 on Transfer Marker between Hands.

\textbf{Operator-time efficiency.}
Demonstrations require continuous teleoperation, DAgger requires rollout monitoring and occasional takeover, whereas RL uses autonomous rollouts with terminal outcome labelling.
Table~\ref{tab:worker-hours} reports the estimated operator time required to collect 100 episodes with each method, including manual resets and rollout monitoring.

\begin{table}[b]
\centering
\caption{Estimated human labor required to collect 100 episodes with SFT, DAgger, and online RL.
The estimates use the normalized collection rates described in the text and should be interpreted as approximate operator-time comparisons rather than direct wall-clock measurements.}
\label{tab:worker-hours}
\begin{tabular}{@{}lc@{}}
\toprule
Collection method & Approx. worker-hours \\
\midrule
SFT demonstrations & 2.0 \\
DAgger corrections & 3.9 \\
Online RL rollouts & 1.2 \\
\bottomrule
\end{tabular}
\end{table}

Using the full pipeline, we obtain 100\%-success policies for Transfer Marker between Hands and Rotate Marker in Hand with 9.7 and 5.2 worker-hours of data collection, respectively.
Under the same collection-time budgets, the corresponding no-RL baselines reach only approximately 75\% with SFT+DAgger and 55\% with SFT, respectively.
Taken together, these results show a favourable speed--performance trade-off for online latent residual RL: it requires only 1.2 worker-hours per 100 rollout episodes, yet reaches higher final success under the same collection-time budget than imitation-only post-training.
Thus, online RL is both more operator-time efficient and more effective at closing the remaining performance gap, rather than merely reducing the cost of data collection.

\subsection{Ablations}
% \label{sec:exp:ablation}

\medskip
\noindent\textit{1) How much does each component contribute?}\par

\begin{figure}[t]
\centering
\includegraphics[width=\columnwidth]{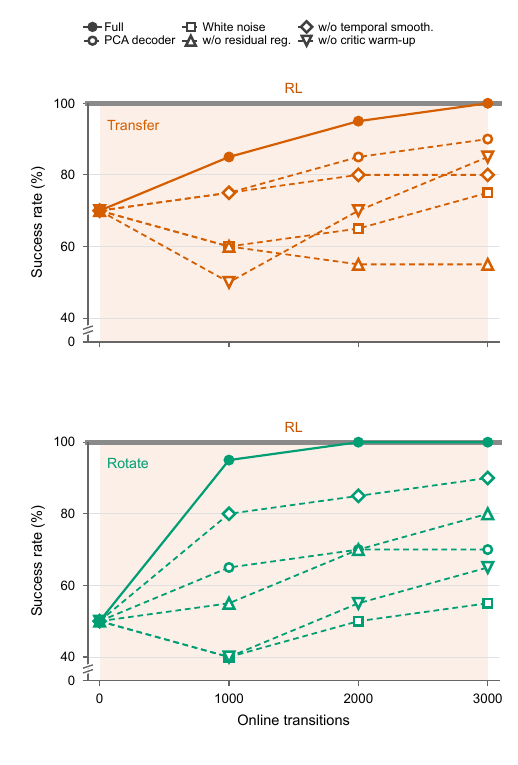}
\caption{Ablation of five latent residual RL design choices on Transfer Marker between Hands and Rotate Marker in Hand.
Each variant changes one component while retaining the remaining optimisation settings.
Curves report measured success rates.}
\label{fig:rl-ablation}
\end{figure}

We ablate five design choices on both online RL tasks, changing one at a time while keeping the rollout budget, safety limits, and optimisation settings fixed.
Figure~\ref{fig:rl-ablation} compares the full method with a frame-wise 9-D PCA representation, temporally independent white exploration noise, removal of the residual regulariser $\mathcal{R}_{\Delta z}$, removal of the temporal smoothness penalty, and removal of critic warm-up.
For the PCA variant, both compression and reconstruction operate independently on each frame, without temporal modelling or conditioning on the initial hand state.
We retrain the reference policy in the PCA action space before freezing it for online residual RL.
The residual actor retains its chunk-wise formulation, exploration noise, and regularisation; the absence of temporal modelling applies only to the PCA representation.

All variants underperform the full method.
The PCA result favours the learned chunk-level representation over frame-wise linear compression under the reported settings.
The representations differ in nonlinearity, temporal context, and initial-hand-state conditioning, and the PCA variant uses a retrained reference policy.
Using temporally independent white noise weakens exploration because its frame-to-frame perturbations do not persist as coherent chunk-level corrections.
Without the residual regulariser $\mathcal{R}_{\Delta z}$, the actor can deviate further from the reference latent action, potentially destabilising learning.
Removing the temporal smoothness penalty can yield less smooth latent action sequences and reduce control quality.
Without critic warm-up, actor updates rely on potentially unreliable early value estimates, which can slow and destabilise online improvement.

\section{Limitations and Discussion}
\label{sec:limitations}

We observe that the effectiveness of residual RL depends strongly on the competence of the frozen reference policy.
When the SFT or DAgger policy succeeds in fewer than approximately 20\% of evaluation trials, residual RL yields little measurable improvement, whereas it becomes consistently useful once the base-policy success rate exceeds approximately 50\%.
This observation suggests that the reference policy must already visit task-relevant states often enough for local residual exploration to obtain informative experience.
We have not systematically characterized the intermediate regime or established that 50\% is a universal threshold across tasks, objects, or embodiments.
Residual RL should therefore be viewed as a late-stage improvement mechanism rather than a substitute for obtaining a minimally competent imitation policy.

The hand-latent dimensionality is also partly constrained by the action interface of the pretrained $\pi_{0.5}$ checkpoint rather than selected solely for optimal RL performance.
Its 32-dimensional action interface allocates 14 dimensions to the two 7-DoF arms, leaving 18 dimensions for the two hands and therefore at most nine latent dimensions per hand under our symmetric layout.
Although the codec sweep characterizes reconstruction quality at several latent widths, we have not decoupled this interface constraint from the effect of latent dimensionality on online RL.
Consequently, our results do not establish that nine dimensions per hand is optimal for other foundation policies, embodiments, or interaction budgets; evaluating latent residual learning with independently variable action capacity remains an important direction for future work.

Finally, the pipeline is designed to improve a specified atomic dexterous skill that completes within approximately $30$\,s.
The experiments do not address long-horizon skill composition or scene-level generalization, and the reported evaluations use local variations around the data-collection conditions rather than substantial changes in lighting, objects, or initial-state distributions.
Extending the method from reliable task-specific post-training to policies that retain robustness across broader scenes and compose multiple dexterous skills remains future work.

\section{Conclusion}
\label{sec:conclusion}

We presented a staged post-training pipeline for adapting pretrained vision--language--action policies to real-world dexterous manipulation.
The pipeline combines supervised fine-tuning, corrective data aggregation, and off-policy residual reinforcement learning through a shared hand-action representation learned from coordinated motion.
Its temporally windowed takeover procedure enables DAgger corrections while retaining absolute hand commands, and its latent residual policy restricts online exploration to a compact space of coordinated hand actions.

Across five marker-manipulation tasks, the successive post-training stages produced final policies that succeeded in all 20 evaluation trials for every task.
Compressing each 20-dimensional hand command to a 9-dimensional latent preserved imitation-learning performance, while latent residual RL improved Transfer Marker between Hands from 70\% to 100\% and Rotate Marker in Hand from 50\% to 100\% within 3,000 real-world transitions.
Under the same interaction budget, residual search in the raw joint space did not obtain the same improvements.
These results support learned action representations as a practical interface between imitation learning and sample-efficient online improvement for high-degree-of-freedom hands.
See the videos on the \href{https://wuji.tech/blog/post-training-0}{project website} for the resulting real-world behaviors.

\section*{ORCID iDs}

\noindent Junlei Zhu: \href{https://orcid.org/0009-0006-6518-3575}{0009-0006-6518-3575}\\
\noindent Shenzhe Yao: \href{https://orcid.org/0009-0006-0016-5501}{0009-0006-0016-5501}\\
\noindent Chaogui Huang: \href{https://orcid.org/0009-0007-3097-1667}{0009-0007-3097-1667}\\
\noindent Wenkai Zhu: \href{https://orcid.org/0009-0002-1928-5934}{0009-0002-1928-5934}\\
\noindent Jingwei Peng: \href{https://orcid.org/0009-0006-9428-6004}{0009-0006-9428-6004}\\
\noindent Guanqi He: \href{https://orcid.org/0009-0007-6016-9800}{0009-0007-6016-9800}\\
\noindent S{\"o}ren Schwertfeger: \href{https://orcid.org/0000-0003-2879-1636}{0000-0003-2879-1636}\\
\noindent Jiahao Chen: \href{https://orcid.org/0000-0002-8927-5646}{0000-0002-8927-5646}\\
\noindent Yide Liu: \href{https://orcid.org/0000-0002-2447-3107}{0000-0002-2447-3107}

\section*{Declaration of conflicting interests}

The authors declare that they have no potential conflicts of interest with respect to the research, authorship, and/or publication of this article.

\section*{Supplemental material}

Supplemental material for this article is available online

% The IJRR bibliography lives beside this source for reliable Overleaf builds.
\bibliographystyle{SageH}
\bibliography{references}

\clearpage
\appendix
% Keep appendix-wide tables at the top of float-only pages instead of vertically
% centring them, while the surrounding appendix text retains the main two-column layout.
\makeatletter
\setlength{\@fptop}{0pt}
\setlength{\@fpbot}{0pt plus 1fil}
\setlength{\@dblfptop}{0pt}
\setlength{\@dblfpbot}{0pt plus 1fil}
\makeatother

\section{Architecture and parameter budget}
\label{app:models}

Table~\ref{tab:model_zoo} lists every module the system runs, its architecture and its exact parameter count, together with whether it is trained, frozen or discarded at each stage.

\begin{table*}[t]
\centering
\caption{Architecture and parameter budget of every module in our system.
Parameter counts are exact, obtained by traversing the Flax parameter tree of
each module under its deployed configuration. The VLA operates on a $32$-D
latent action space ($2\times7$ arm joints $+$ $2\times9$ hand latents) with a
prediction horizon $H{=}32$; the residual RL policy acts on chunks of length
$C{=}25$.}
\label{tab:model_zoo}
\footnotesize
\setlength{\tabcolsep}{4pt}
\renewcommand{\arraystretch}{1.15}
\newcommand{\grouprow}[1]{\multicolumn{5}{@{}l}{\textit{#1}} \\ \addlinespace[1pt]}
\begin{tabular}{@{}l
  >{\raggedright\arraybackslash}p{3.3cm}
  >{\raggedright\arraybackslash}p{3.6cm} r l@{}}
\toprule
\textbf{Component} & \textbf{Architecture} & \textbf{Key configuration} &
\textbf{\#Params} & \textbf{Status} \\
\midrule
\grouprow{(a) Hand latent codec: per-hand chunk VAE, $20$-D $\rightarrow 9$-D per step}
Encoder (per hand) & Time-token Transformer (bidir.) & $d{=}128$, $L{=}3$, $h{=}4$, MLP$\times2$ & 0.43\,M & frozen \\
Decoder (per hand) & Time-token Transformer (bidir.) & $d{=}128$, $L{=}3$, $h{=}4$, MLP$\times2$ & 0.43\,M & frozen \\
\multicolumn{3}{@{}l}{\textit{Subtotal}, left $+$ right, independent weights} & \textbf{1.70\,M} & \\
\midrule
\grouprow{(b) Base VLA: $\pi_{0.5}$ fine-tuned to emit native $32$-D latent actions}
Vision encoder & SigLIP-So400m/14 ViT & $d{=}1152$, $L{=}27$, $h{=}16$, $224^2$, 3 cams & 414.80\,M & frozen$^\dagger$ \\
VLM backbone & Gemma-2B (MQA) & $d{=}2048$, $L{=}18$, $h{=}8$/$1$ KV, $d_{\text{ff}}{=}16384$, $N_{\mathrm{vocab}}{=}257{,}152$ & 2.51\,B & frozen$^\dagger$ \\
Action expert & Gemma-300M $+$ adaRMSNorm & $d{=}1024$, $L{=}18$, $h{=}8$/$1$ KV, $d_{\text{ff}}{=}4096$ & 427.93\,M & frozen$^\dagger$ \\
Flow I/O heads & Linear (action in/out, time MLP) & $32\!\rightarrow\!1024$, $1024\!\rightarrow\!32$ & 2.17\,M & frozen$^\dagger$ \\
\multicolumn{3}{@{}l}{\textit{Subtotal}} & \textbf{3.35\,B} & \\
\midrule
\grouprow{(c) RL token: prefix $[M,2048] \rightarrow h_{\mathrm{rl}}\in\mathbb{R}^{2048}$}
Readout encoder & 2$\times$(self-attn $+$ cross-attn $+$ GeGLU MLP) & $d{=}2048$, $h{=}8$, MLP$\times4$ & 406.96\,M & frozen \\
AR decoder & 2$\times$(causal self-attn $+$ cross-attn $+$ GeGLU) & $d{=}2048$, $h{=}8$, MLP$\times4$ & 415.35\,M & train-only \\
\multicolumn{3}{@{}l}{\textit{Subtotal}} & \textbf{822.32\,M} & \\
\midrule
\grouprow{(d) Residual RL policy: chunk-level TD3 on latent chunks}
ChunkActor & 3 proj. $+$ 2-layer MLP (Dense-LN-swish) & $d_h{=}256$, zero-init $\tanh$ head & 1.21\,M & trainable \\
TwinCritic & 2 independent $Q$ networks, same trunk & $d_h{=}256$, $1.01$\,M each & 2.02\,M & trainable \\
\multicolumn{3}{@{}l}{\textit{Subtotal}$^\ddagger$} & \textbf{3.23\,M} & \\
\midrule
\multicolumn{3}{@{}l}{\textbf{Total (all modules)}} & \textbf{4.18\,B} & \\
\multicolumn{3}{@{}l}{\textbf{Deployed at inference}: VLA $+$ hand decoders $+$ RL-token encoder $+$ actor} & \textbf{3.76\,B} & \\
\multicolumn{3}{@{}l}{\textbf{Trainable during online RL}} & \textbf{3.23\,M} & $0.08\%$ \\
\bottomrule
\end{tabular}
\vspace{2pt}
\begin{flushleft}\footnotesize
$^\dagger$ The full $3.35$\,B VLA is fine-tuned end-to-end during latent SFT and each DAgger round
(30\,k updates per phase, batch $64$, cosine $5\!\times\!10^{-5}$) and kept \emph{frozen}
during RL-token training and online RL.\\
$^\ddagger$ Target copies add another $3.23$\,M during training.
\end{flushleft}
\end{table*}

\section{Training schedules}
\label{app:schedules}

\subsection{Hand-action codec}
Table~\ref{tab:codec-comparison} summarises the architectures used in the hand-representation comparison of Figure~\ref{fig:codec-analysis}.
Each method uses a nine-dimensional latent representation per frame.

\begin{table*}[t]
\caption{Architectures used in the hand-representation comparison.
Shapes and parameter counts are reported for one hand with $H=32$ and a nine-dimensional latent per frame.
In the chunk-based mappings, the additional 20-dimensional input is the measured hand state $q_0$ at the start of the chunk.}
\label{tab:codec-comparison}
\centering
\scriptsize
\setlength{\tabcolsep}{3pt}
\renewcommand{\arraystretch}{1.15}
\begin{tabular}{@{}>{\raggedright\arraybackslash}p{1.55cm}
  >{\raggedright\arraybackslash}p{2.35cm}
  >{\raggedright\arraybackslash}p{3.55cm}
  >{\raggedright\arraybackslash}p{2.15cm}
  >{\raggedright\arraybackslash}p{2.15cm}@{}}
\toprule
\textbf{Method} & \textbf{Shape} &
\textbf{Encoder / decoder} & \shortstack[l]{\textbf{Parameters}\\\textbf{(enc. / dec. / total)}} &
\textbf{Embedding} \\
\midrule
Frame PCA & $20 \rightarrow 9 \rightarrow 20$ & Nine principal components & --- & None \\
Frame MLP & $20 \rightarrow 9 \rightarrow 20$ & MLP widths $(128,64)$ / $(64,128)$ & 11.9\,K / 11.9\,K / 23.8\,K & None \\
Frame Transformer & $20 \rightarrow 9 \rightarrow 20$ & 2 layers per side, width 64, 4 heads, MLP$\times2$ & 68.9\,K / 68.3\,K / 137.2\,K & Finger positions; decoder queries \\
Chunk MLP & $\begin{array}{c}(H{\times}20,20)\\[-2pt]\rightarrow H{\times}9\\[-2pt]\rightarrow H{\times}20\end{array}$ & Flattened chunk; 2 hidden layers per side, width 512 & 0.90\,M / 0.75\,M / 1.65\,M & None \\
Chunk Transformer & $\begin{array}{c}(H{\times}20,20)\\[-2pt]\rightarrow H{\times}9\\[-2pt]\rightarrow H{\times}20\end{array}$ & 3 layers per side, width 128, 4 heads, MLP$\times2$ & 426.3\,K / 425.1\,K / 851.4\,K & Temporal positions; $q_0$ token \\
\bottomrule
\end{tabular}
\end{table*}

Separate models are fitted for the left and right hands.
All methods use the same episode-level 90/10 training--validation split with seed 0, with every frame from an episode assigned to the same partition; failed episodes are retained.
The same per-joint standardisation procedure is applied using statistics computed from the training split only.
Neural checkpoints are selected by the lowest validation per-joint MAE in the original joint units.
PCA is fitted to the standardised training frames using nine components and evaluated on the same held-out episodes.
The projection and reconstruction are applied independently to each frame, using a fixed basis for each hand.
PCA uses neither temporal context nor conditioning on the initial hand state $q_0$, and is fitted without a temporal training objective.

The frame-based neural models are trained for 200 epochs with AdamW, a batch size of 4096, cosine learning-rate decay from $10^{-3}$, and weight decay $10^{-4}$.
They use the same weighted Huber reconstruction loss; the frame Transformer additionally uses a KL weight of $10^{-3}$, linearly warmed up over the first 2000 steps.
The chunk-based models are trained for 30\,k steps with AdamW, a batch size of 4096, cosine decay from $3\!\times\!10^{-4}$, and weight decay $10^{-4}$.
In \eqref{eq:codec-loss}, the Huber threshold is $\tau_{\mathrm{Huber}}=1$; the thumb, index, and middle-finger joints have weight $1$, while the ring- and little-finger joints have weight $0.5$; and $(\lambda_0,\lambda_1,\lambda_2)=(1,0.2,0.05)$.
The KL weight $\beta_{\mathrm{KL}}$ is zero for the first 4000 steps, increases linearly to $3\!\times\!10^{-4}$ over the next 4000 steps, and remains constant thereafter.

\subsection{Reference-policy training}
Table~\ref{tab:sft-schedule} gives the training settings for the SFT and DAgger checkpoints in Figure~\ref{fig:sft}.
Within each stage, these settings are shared across tasks and dataset sizes; the latent and raw variants differ in the space used for flow matching, and the raw variant omits the decoded channel of \eqref{eq:sft-loss}, which has no meaning without the codec.
Only the arm and hand weights are then in force, at $\lambda_{\mathrm{arm}}$ on all $54$ dimensions.

The aggregation runs of Figure~\ref{fig:sft} reuse the same optimisation schedule.
After all 50 corrections in a round have been collected, the VLA is reinitialised from the same public $\pi_{0.5}$ checkpoint and trained for 30\,k update steps; a fraction $\alpha_{\mathrm{new}}$ of each minibatch is drawn from the newly collected corrections and the remainder from the previously accumulated demonstrations and corrections.
We use $\alpha_{\mathrm{new}}=0.5$ in all aggregation runs.

\begin{table*}[t]
\caption{Full training schedule for the reference policy.
Each SFT policy and each DAgger round is trained for 30\,k updates from the same public $\pi_{0.5}$ checkpoint.}
\label{tab:sft-schedule}
\centering
\footnotesize
\setlength{\tabcolsep}{4pt}
\renewcommand{\arraystretch}{1.12}
\begin{tabular}{@{}p{2.15cm} >{\raggedright\arraybackslash}p{4.3cm} p{2.15cm} >{\raggedright\arraybackslash}p{4.3cm}@{}}
\toprule
\textbf{Setting} & \textbf{Value} & \textbf{Setting} & \textbf{Value} \\
\midrule
\multicolumn{4}{@{}l}{\textit{Initialisation and objective}} \\
Initialisation & same public $\pi_{0.5}$ checkpoint for every SFT policy and DAgger round & Updated parameters & all $3.35$\,B (vision encoder, backbone, action expert, flow I/O heads) \\
Objective & \eqref{eq:sft-loss}, conditional flow matching plus decoded hand channel & Loss weights & $\lambda_{\mathrm{arm}} {=} 1.0$, $\lambda_{\mathrm{lat}} {=} 0.25$, $\lambda_{\mathrm{dec}} {=} 0.75$ \\
Flow time $t_{\mathrm{flow}}$ & $0.999\,\operatorname{Beta}(1.5,1)+0.001$ & & \\
\midrule
\multicolumn{4}{@{}l}{\textit{Optimisation}} \\
Steps & $30$\,k per SFT policy or DAgger round & Batch size & $64$ chunks \\
Optimiser & AdamW, $\beta_1 {=} 0.9$, $\beta_2 {=} 0.95$, $\epsilon {=} 10^{-8}$ & Weight decay & $10^{-10}$ \\
Gradient clipping & global norm $1.0$ & Warmup & linear, $1000$ steps \\
Decay & cosine over $30$\,k steps, peak $5 \times 10^{-5}$, floor $5 \times 10^{-6}$ & Precision & bfloat16 parameters and activations, float32 optimiser state \\
EMA & decay $0.99$ & Hardware & 8$\times$H20 \\
Wall-clock & 12\,h per run & & \\
\midrule
\multicolumn{4}{@{}l}{\textit{Data and inference}} \\
Views & 3 cameras at $224^2$ & Prompt/state token IDs & $T_{\mathrm{prompt}}^{\max}{=}256$ (task, template, and quantized state; padded or truncated) \\
Chunk horizon $H$ & $32$ & Executed frames $C$ & $25$ \\
Sampling & SFT: uniform over episode start frames; DAgger: $50\%$ new corrections, $50\%$ prior data & Normalisation & per dimension, statistics from the training split only \\
Image augmentation & non-wrist views: 95\% random crop and $\pm5^\circ$ rotation; all views: color jitter ($0.3/0.4/0.5$ brightness/contrast/saturation) & Integration & $10$ Euler steps of the learned field \\
Decoding & hand latents through the frozen $D^{\mathrm{hand}}$, arm blocks by identity & & \\
\bottomrule
\end{tabular}
\end{table*}

\clearpage
% Restore full column height after flushing the preceding double-column tables.
\makeatletter
\global\@colht\textheight
\global\@colroom\textheight
\global\vsize\textheight
\@floatplacement
\makeatother
\raggedbottom
\subsection{Offline RL-token pretraining}
\label{app:rlt-pretraining}

\begin{table}[!t]
\caption{Default training configuration for offline RL-token pretraining.
Optimiser, learning-rate, and EMA settings follow the base-policy defaults.}
\label{tab:rlt-schedule}
\centering
\footnotesize
\setlength{\tabcolsep}{4pt}
\renewcommand{\arraystretch}{1.12}
\begin{tabular}{@{}>{\raggedright\arraybackslash}p{0.39\columnwidth} >{\raggedright\arraybackslash}p{\dimexpr0.61\columnwidth-2\tabcolsep\relax}@{}}
\toprule
\textbf{Setting} & \textbf{Value} \\
\midrule
\multicolumn{2}{@{}l}{\textit{Optimisation}} \\
Optimiser & AdamW \\
AdamW parameters & $\beta_1 {=} 0.9$, $\beta_2 {=} 0.95$, $\epsilon {=} 10^{-8}$ \\
Weight decay & $10^{-10}$ \\
Gradient clipping & global norm $1.0$ \\
EMA decay & $0.99$ \\
Network dropout & $0$ \\
\midrule
\multicolumn{2}{@{}l}{\textit{Learning-rate schedule}} \\
Warmup & linear, $1000$ updates \\
Peak learning rate & $5 \times 10^{-5}$ \\
Cosine decay endpoint & $30$\,k updates \\
Final learning rate & $5 \times 10^{-6}$, held from $30$\,k to $40$\,k updates \\
\midrule
\multicolumn{2}{@{}l}{\textit{Training setup}} \\
Training budget & $40$\,k updates \\
Global batch size & $64$ observations \\
Training seed & $0$ \\
Parallelism & fully sharded data parallelism across eight devices \\
\bottomrule
\end{tabular}
\end{table}

The RL-token readout is pretrained on observations from the reference-policy demonstration set using the reconstruction objective in \eqref{eq:readout-loss}.
Observation preprocessing is the same as for the reference policy.
The VLA remains frozen; only the readout token, encoder, output projection, and auxiliary decoder are optimised.
Neither action labels nor rewards enter this objective.

Table~\ref{tab:model_zoo} specifies the encoder and decoder architectures.
Within each encoder block, the second attention sublayer takes its keys and values from the input to that block.
The decoder is trained with teacher forcing, using right-shifted target features preceded by a learned beginning-of-sequence embedding and the RL token as cross-attention memory.
Padded image-token positions are excluded from the encoder's attention keys and the reconstruction loss.
The minibatch loss is averaged over all valid image-token positions and feature dimensions.

Table~\ref{tab:rlt-schedule} summarises the default offline training configuration, including the optimiser, learning-rate schedule, and exponential moving average (EMA) settings inherited from the base policy.
The learning rate follows linear warmup and cosine decay, then remains at its minimum for the final 10\,k updates.
Training uses a fixed update budget, without validation-based checkpoint selection or early stopping.
Checkpoints are saved every 5000 updates and at the end of the 40\,k-update schedule.
The default configuration uses EMA weights for inference.
After pretraining, the auxiliary decoder is discarded and the readout is frozen for online RL.

\subsection{Online residual reinforcement learning}
Table~\ref{tab:rl-schedule} gives the online training and collection settings after the reference policy and RL-token readout have been frozen.

\begin{table*}[t]
\caption{Training and collection schedule for latent residual RL.
The same optimisation settings are used for Transfer Marker between Hands and Rotate Marker in Hand; only the task-specific residual mask differs.}
\label{tab:rl-schedule}
\centering
\fontsize{7.5}{8.5}\selectfont
\setlength{\tabcolsep}{4pt}
\renewcommand{\arraystretch}{1.15}
\newcommand{\grouprowrl}[1]{\multicolumn{2}{@{}l}{\textit{#1}} \\ \addlinespace[1pt]}
\begin{tabular}{@{}>{\raggedright\arraybackslash}p{0.30\textwidth} >{\raggedright\arraybackslash}p{0.64\textwidth}@{}}
\toprule
\textbf{Setting} & \textbf{Value} \\
\midrule
\grouprowrl{Objective and updates}
Algorithm & TD3 \\
Actor objective & Equation~\ref{eq:actor-loss} \\
Loss weights & $\beta_q {=} 0.5$, $\beta_{\Delta z} {=} 3.0$, $\beta_s {=} 0.5$ \\
Actor learning rate & $1 \times 10^{-4}$ \\
Critic learning rate & $5 \times 10^{-5}$ \\
Optimisers & Adam for actor and critics; no weight decay \\
Batch size & $128$ transitions \\
Discount & per-frame $\gamma {=} 0.99$; chunk discount $0.99^{25}$ \\
Target update & Polyak coefficient $\tau_{\mathrm{Polyak}} {=} 5 \times 10^{-3}$ \\
Actor update frequency & one actor update per critic update during joint training \\
Update cadence & $200$ gradient steps after each newly collected episode \\
Replay capacity & $200$\,k transitions \\
Replay sampling & recency weighted $6{:}5{:}4{:}3{:}2$ over the five latest episodes; latest episode guaranteed in each batch \\
\midrule
\grouprowrl{Action and interaction}
Reference policy & frozen after SFT/DAgger \\
Chunk horizon & $H {=} 32$ \\
Executed frames & $C {=} 25$ at 30\,Hz \\
Residual scope & task-specific mask, Equation~\ref{eq:task-residual-mask} \\
Actor initialisation & zero output layer \\
Reference dropout & probability $0.5$ per training sample \\
Residual bounds & arm increment $0.002$ per frame; hand residual $0.10$ in latent coordinates \\
Execution rate limit & maximum joint-position change $\pi/18\,\mathrm{rad}$ per control frame \\
Warm-up & critics only after $500$ replay transitions; actor enabled after manual stability check \\
Online budget & 3000 transitions per task \\
Reward & terminal failure $-1$; success and non-terminal chunks $0$ \\
\midrule
\grouprowrl{Exploration and evaluation}
Arm exploration & white noise on increments, accumulated across the chunk \\
Hand exploration & direct per-dimension Brownian noise ($\nu{=}2$, $1/f^2$) \\
Noise scales & $\sigma^{\mathrm{arm}} {=} 0.01$, $\sigma^{\mathrm{hand}} {=} 0.05$ \\
Evaluation noise & none \\
Evaluation frequency & every 1000 transitions \\
Evaluation trials & 20 per checkpoint \\
\bottomrule
\end{tabular}
\end{table*}

\textbf{Matched residual exploration.}
For the raw-space comparison of Section~\ref{sec:exp:rl}, we keep the TD3 updates, safety limits, and temporal structure of the exploration noise unchanged.
Both residual methods use white noise on the arm increments, accumulated across the chunk, and the same direct per-dimension normalized Brownian noise ($\nu{=}2$, $1/f^2$) on the hand block.
The hand block is in latent coordinates for our method and in joint coordinates for the raw residual.
Their noise scales are matched by the magnitude of the resulting perturbations in joint space after decoding, so the comparison changes the action representation without changing the physical exploration scale.

\end{document}